\documentclass{article}

\usepackage[dvipsnames]{xcolor}
\definecolor{linkcolor}{rgb}{0,0.6,0.6}
\usepackage[colorlinks=true,
    linkcolor=linkcolor,
    citecolor=linkcolor,
    filecolor=linkcolor,
    urlcolor=linkcolor,
    ]{hyperref} 
\hypersetup{
    colorlinks=true,
    linkcolor=linkcolor,
    filecolor=magenta,      
    urlcolor=ForestGreen,
    }
\usepackage{amsfonts}
\usepackage{nicefrac}       %
\usepackage{amsmath}
\usepackage{amssymb}
\usepackage{cancel}
\usepackage{mathtools}
\usepackage{bbm}
\usepackage{bm}

\usepackage{microtype}      %

\usepackage{enumitem}

\newcommand{\bg}[1]{\boldsymbol{#1}}

\newcommand{\vect}[1]{\mathbf{#1}}

\newcommand{\ixP}{\mathcal{P}}

\newcommand{\vz}{\vect{z}}
\newcommand{\vx}{\vect{x}}
\newcommand{\vy}{\vect{y}}
\newcommand{\vu}{\vect{u}}

\newcommand{\vthetat}{\bg{\tilde{\theta}}}

\newcommand{\vtheta}{\bg{\theta}}
\newcommand{\vphi}{\boldsymbol{\phi}}

\newcommand{\lambdapen}{\lambda_{\text{pen}}}
\newcommand{\lambdaco}{\lambda_{\text{co}}}

\newcommand{\gatelr}{\eta_{\text{primal}}^{\vphi}}
\newcommand{\weightlr}{\eta_{\text{primal}}^{\vthetat}}
\newcommand{\duallr}{\eta_{\text{dual}}}

\newcommand{\blobletter}[1]{\raisebox{.5pt}{\textcircled{\raisebox{-.8pt}{{\hspace{-1mm} \small #1}}}}}

\usepackage{microtype}
\usepackage{graphicx}
\usepackage{subfigure}
\usepackage{booktabs} %

\usepackage{amsmath}
\usepackage{amssymb}
\usepackage{mathtools}
\usepackage{amsthm}

\usepackage[capitalize]{cleveref}

\theoremstyle{plain}

\theoremstyle{definition}

\theoremstyle{remark}

\usepackage{multirow}
\usepackage{float}
\floatstyle{plaintop}
\restylefloat{table}
\usepackage[tableposition=top]{caption}

\usepackage[symbol]{footmisc}

\usepackage[textsize=tiny, colorinlistoftodos]{todonotes}

\newcounter{todocounter}

\usepackage{tikz}
\usepackage{pgfplots}
\pgfplotsset{compat=1.17}
\usetikzlibrary{decorations.markings}

\usepackage{scrwfile}
\TOCclone[\appendixname]{toc}{atoc}

\AfterTOCHead[toc]{%
}
\AfterTOCHead[atoc]{%
  \edef\maintocdepth{\the\value{tocdepth}}%
  \value{tocdepth}=-10000\relax%
}

\PassOptionsToPackage{numbers}{natbib}
     \usepackage[final]{neurips2022}

\usepackage[utf8]{inputenc} %
\usepackage[T1]{fontenc}    %
\usepackage{hyperref}       %
\usepackage{url}            %
\usepackage{booktabs}       %
\usepackage{amsfonts}       %
\usepackage{nicefrac}       %
\usepackage{microtype}      %
\usepackage{xcolor}         %

\definecolor{myblue}{HTML}{0035d5}
\definecolor{myred}{HTML}{ce2727}
\definecolor{myorange}{HTML}{ff8c42}
\definecolor{mygreen}{HTML}{9dbf98}

\newcommand{\loonie}{{\color{myblue} L$_0$onie}}
\newcommand{\coin}{{\color{myorange} COIN}}
\newcommand{\mprune}{{\color{myred} magnitude pruning}}
\newcommand{\jpeg}{{\color{mygreen} JPEG}}

\title{L$_0$onie: Compressing COINs with L$_0$-constraints}

\author{
   \textbf{Juan Ramirez} 
   \hspace{8mm}  
   \textbf{Jose Gallego-Posada} \\
   \texttt{\{juan.ramirez, gallegoj\}@mila.quebec} \\
   Mila and DIRO, Université de Montréal, Canada
}

\def \FigurePath{workshop_figs}

\begin{document}

\maketitle

\vspace{-2ex}

\begin{abstract}
\vspace{-1ex}
Advances in Implicit Neural Representations (INR) have motivated research on domain-agnostic compression techniques. These methods train a neural network to approximate an object, and then store the weights of the trained model. For example, given an image, a network is trained to learn the mapping from pixel locations to RGB values.
In this paper, we propose L$_0$onie, a sparsity-constrained extension of the COIN compression method. Sparsity allows to leverage the faster learning of overparameterized networks, while retaining the desirable compression rate of smaller models. Moreover, our constrained formulation ensures that the final model respects a pre-determined compression rate, dispensing of the need for expensive architecture search.
\end{abstract}

\vspace{-3ex}

\section{Introduction}
\label{sec:introduction}

Implicit Neural Representations (INRs) train neural networks mapping coordinates (e.g. pixel locations) to features (e.g. RGB values) in order to approximate a given object. INRs have been applied to a wide range of data modalities including audio \citep{sitzmann2020implicit}, images \citep{stanley2007compositional}, video \citep{li2021neural}, 3D scenes \citep{mescheder2019occupancy} and temperature fields \citep{dupont2021generative}. INRs provide a new perspective on data compression: rather than dealing with the ``raw'' features of the object, train an INR to approximate the object and store the parameters of the learned model. COIN \citep{dupont2021coin} pioneered this approach for image compression.

INR-based compression is a nascent technology. COIN exhibits sub-par performance and more intensive computational cost compared to domain-specific codecs. Moreover, it commonly requires architecture search to balance the relationship between the model capacity and reconstruction quality. Although these techniques are not yet competitive with established codecs for well-studied domains like audio or image compression, the greater promise of methods like COIN lies on their ability to provide compression standards that can be applicable to virtually any data modality.

Recent works \citep{dupont2022coinpp, lee2021meta, schwarz2022meta} focus on improving COIN via a meta-learning approach in which a base network is (pre-)trained over a large collection of datapoints, so that instance-dependent INRs can be found quickly. The task of encoding of individual instances is casted as a search for \textit{modulations} \citep{perez2017film} on the base architecture.

\begin{figure}
\centering
\begin{minipage}{.55\textwidth}
    \hspace{-15mm}
    \begin{tabular}{ccccc}
    & & 1.2 BPP & 0.6 BPP &  0.3 BPP\\
    Original  & \rotatebox{90}{ \hspace{3mm} \textbf{\loonie}} & \includegraphics[width=0.23\textwidth,trim={0mm 0mm 0mm 0mm},clip]{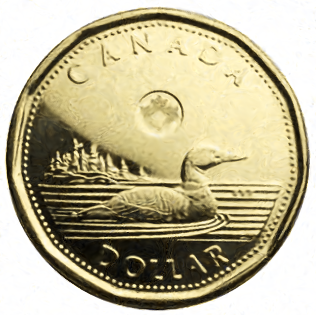} &  \includegraphics[width=0.23\textwidth,trim={0mm 0mm 0mm 0mm},clip]{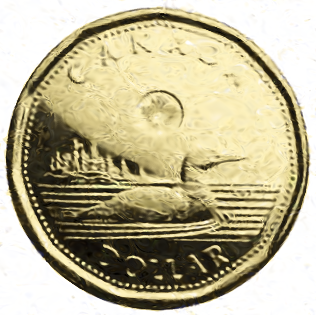} & 
    \includegraphics[width=0.23\textwidth,trim={0mm 0mm 0mm 0mm},clip]{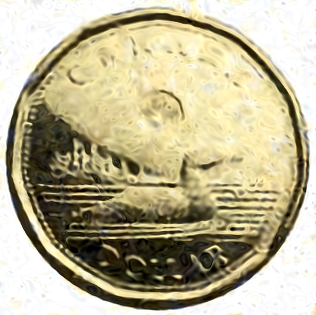} \\
    \includegraphics[width=0.23\textwidth,trim={0mm 0mm 0mm 0mm},clip]{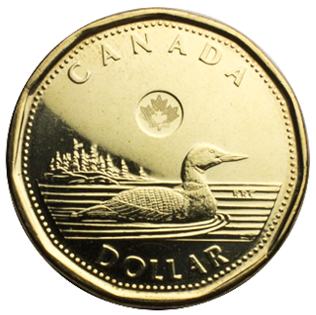} & \rotatebox{90}{ \hspace{3mm} \textbf{\coin}} & \includegraphics[width=0.23\textwidth,trim={0mm 0mm 0mm 0mm},clip]{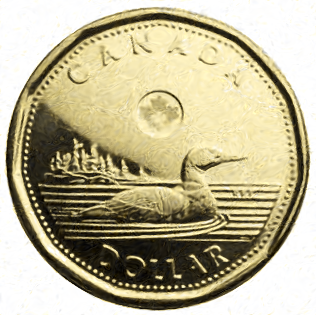} &  \includegraphics[width=0.23\textwidth,trim={0mm 0mm 0mm 0mm},clip]{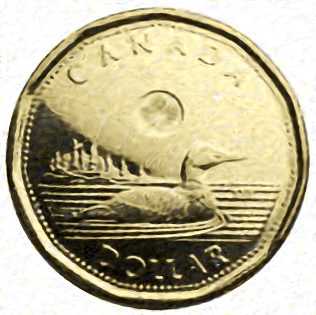} & 
    \includegraphics[width=0.23\textwidth,trim={0mm 0mm 0mm 0mm},clip]{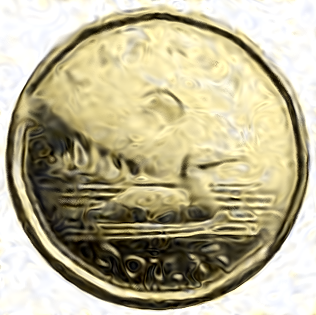} \\& \rotatebox{90}{ \hspace{3mm} \textbf{\jpeg}} & \includegraphics[width=0.23\textwidth,trim={0mm 0mm 0mm 0mm},clip]{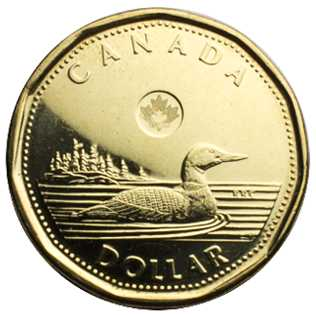} &  \includegraphics[width=0.23\textwidth,trim={0mm 0mm 0mm 0mm},clip]{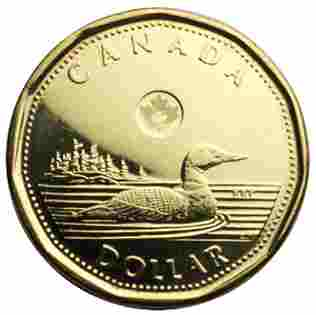} & 
    \includegraphics[width=0.23\textwidth,trim={0mm 0mm 0mm 0mm},clip]{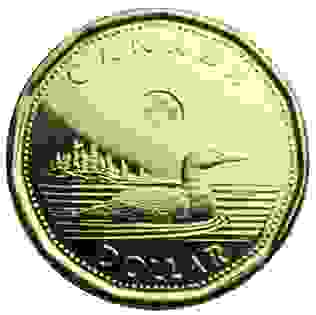}
    \end{tabular}
    \captionof{figure}{Qualitative comparison between \loonie, \coin~and \jpeg~on a picture of a loonie.}
    \label{fig:loonie_comparison}
\end{minipage}%
\hfill
\begin{minipage}{.4\textwidth}
    \centering
    \vspace{7mm}
    \includegraphics[width=1.\textwidth,trim={2mm 2mm 2mm 2mm},clip]{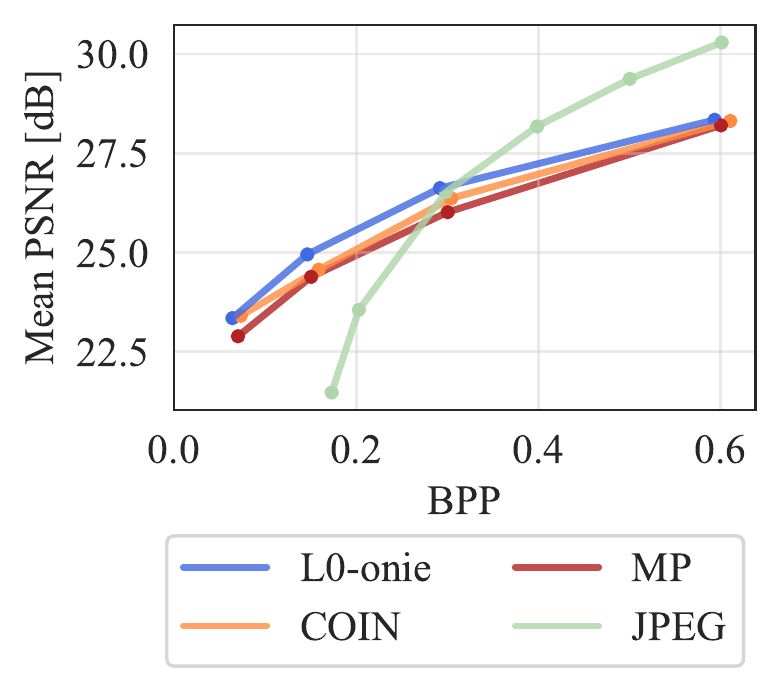}
    \vspace{.5mm}
    \captionof{figure}{PSNR achieved by different techniques over the entire Kodak dataset.} 
    \label{fig:psnr_vs_bpp}
\end{minipage}
\end{figure}

In this work we concentrate on the interplay between the model size, its representational capacity and the required training time. In L$_0$onie\footnote{Loonie is a colloquial name for the Canadian one dollar \textit{coin}.}, we combine ``overparameterized'' COIN models with the sparse L$_0$-reparametrization of \citet{louizos2017learning}. This allows us to exploit the power of larger models to achieve better reconstructions, faster; without having to commit to their undesirable compression rate. We take advantage of the inherent redundancies in these larger models, and sparsify them during training in order to achieve a pre-specified compression rate. Our constrained formulation provides direct control over the resulting compression rate and removes the need for costly hyper-parameter tuning or architecture search. \cref{fig:loonie_comparison} provides an overview of the compression behavior of of L$_0$onie, COIN and JPEG at different BPP budgets. 

Our code is available at: \texttt{\url{https://github.com/juan43ramirez/l0onie}}.

\section{Implicit Neural Representations}
\label{sec:inr}

For simplicity, we present a description of INRs using terminology from image processing. However, we highlight that INRs are in principle agnostic\footnote{Note that a choice of model architecture and activation functions might not be suitable for all data modalities.} to the ``data type'' at hand. Consider the image $\mathbf{I}$ to be compressed as a collection of pixel coordinates $\vx_p \in [-1, 1]^2$ and corresponding RGB pixel intensities $\vy_p \in [0, 1]^3$, indexed over a discrete set of pixels $\ixP$. 

Consider a family of neural networks $f_{\vtheta}: \mathbb{R}^2 \rightarrow \mathbb{R}^3$ with parameters $\vtheta$. An implicit neural representation (INR) of image $\mathbf{I}$ is a solution to the following supervised learning problem: 
\begin{equation}
    \label{eq:inr}
    \min_{\vtheta} \mathcal{L}_{\mathbf{I}}(\vtheta) \triangleq \sum_{p \in \ixP} || f_{\vtheta}(\vx_p) - \vy_p||^2_2.
\end{equation}
Upon solving the optimization problem, the original image can be (approximately) reconstructed by \textit{evaluating} the function $f_{\vtheta}$ at the pixel locations $\ixP$. The choice of a family of models can have a large influence in the reconstruction performance, and is an active area of research. Recent proposals combine fully-connected models with positional embeddings \citep{tancik2020FourierFeatures} or sine activation functions \citep{sitzmann2020implicit}. 

\citet{dupont2021coin} propose COIN as a creative approach for image compression: when compressing an image $\mathbf{I}$, rather than storing RGB values, store the \textit{weights} of an INR of $\mathbf{I}$. We follow COIN's choice of MLPs with sinusoidal activation functions in this work. 

The distortion-rate tradeoff in COIN involves balancing the higher expressive capacity of large models with the detrimental effect of more model parameters towards the compression rate. Achieving specific compression rates while retaining low distortion may involve time-consuming architecture search. 

\subsection{Sparsifying INRs}
\label{sec:sparse_inr}

Given that compressing images using COIN already requires more time than standard methods like JPEG (by several orders of magnitude), dispensing of the need for architecture search is an important step towards a wider adoption of INR-based compression. Sparsity techniques are a natural approach to address this issue.

Magnitude pruning can be used to attain specific compression rates, but results in sub-optimal performance even after fine-tuning (see \cref{sec:experiments}). The assumptions made by heuristic-based sparsity methods popular in other machine learning tasks might not transfer well to the context of INRs.\footnote{E.g. magnitude pruning relies on the idea that low-norm parameters \textit{should} have a small influence in the final prediction. This has been shown empirically for image classification tasks where the model predicts a label among a discrete number of classes \citep{li2017l1pruning, gale2019state}. We hypothesize that the sub-par performance of magnitude pruning in our experiments may be due to the use of sine activations in the model and the continuous nature of the targets.} 

\citet{louizos2017learning} propose a framework for learning sparse models by means of a differentiable reparametrization of the model weights $\vtheta = \vthetat \odot \vz$, where $\vthetat$ are free (signed) parameter magnitudes, and $\vz$ are stochastic gates indicating whether a parameter is active. The gates follow a \textit{hard-concrete} distribution \citep{louizos2017learning} with parameters $\vphi$. The authors then augment the usual training objective with an additive penalty given by the expected L$_0$-norm of the gates $\vz$, to encourage sparsity in the model.

We propose to combine the MLPs with sinusoidal activations of COIN with this L$_0$ reparametrization. This sparsity perspective opens the door for training ``overparameterized'' models which can achieve better performance, faster \citep{aroraOptimizationDeepNetworks2018a}. Moreover, learning the sparsity pattern during training avoids having to commit to the undesirably low compression rate of the fully dense model.

The same reparametrization is used in the concurrent work of \citet{schwarz2022meta}, although stemming from a different motivation. While the authors provide valuable insights regarding the integration of sparsity within a meta-learning pipeline, they overlook the challenge of tuning the hyper-parameter $\lambdapen$, which dictates the relative importance of the sparsity term. Adjusting $\lambdapen$ to achieve a \textit{specific} compression rate can be as prohibitive as performing architecture search \citep{boyd2004convex, gallego2021flexible}.

Note that the stochastic re-parametrization induces a distribution over the models. For practical reasons, it is convenient to have a single model at decoding time. We use the gate medians $\hat{\vz}(\vphi) \in [0, 1]$ for constructing said model. Due to the ``stretching'' in the hard-concrete distribution, it is possible for the medians to be exactly 0 or 1 and not just fractional. See \cref{app:gates} for more details.

\subsection{L$_0$-constrained formulation}
\label{sec:sparse_inr}

\citet{gallego2021flexible} argue that constrained formulations can provide greater hyper-parameter interpretability and controllability when learning sparse neural networks, compared to the commonly used penalized approach. The authors consider constraints on the expected L$_0$-norm of the parameters using the reparametrization of \citet{louizos2017learning}. We extend their constrained formulation by using \textit{proxy-constraints} \citep{cotter2019} to directly control the compression rate of the resulting model (see \cref{app:proxy_constraints}).

We consider the problem of finding a sparse INR for an image $\mathbf{I}$ given a constraint on the compression rate expressed in terms of a budget of $\tau_{\texttt{BPP}}$ bits-per-pixel. We observed degraded performance when training using Monte Carlo samples for the stochastic gates, as done in \citet{louizos2017learning}. For this reason, we train a deterministic model using the gate medians. This setting coincides with the decoding time model described above.

Formally, we consider the following constrained optimization problem:
\vspace{-1ex}
\begin{align}
    \label{eq:const_inr}
    \underset{\boldsymbol{\tilde{\theta}}, \vphi}{\text{min}} \, \mathcal{L}_{\mathbf{I}}(\vthetat \odot \hat{\vz}(\vphi)) \, \,
    \text{s.t.} \hspace{2mm} \texttt{BPP}(\texttt{cast}(\vthetat \odot \hat{\vz}(\vphi)) \le \tau_{\texttt{BPP}},
\end{align}
where $\texttt{BPP}(\texttt{cast}(\vu)) = \sum_i  \mathbb{I}[\texttt{cast}(\vu_i) \ne 0] \cdot  \texttt{bits}(\texttt{cast}(\vu_i))$, and $\texttt{cast}(\cdot)$ quantizes its input to a specified data type, such as \texttt{float16}.

In practice, we optimize the Lagrangian associated with the problem in \cref{eq:const_inr}. Let $\lambdaco \ge 0$ be the Lagrange multiplier corresponding to the constraint. The min-max Lagrangian problem is given by:
\begin{equation}
    \label{eq:minmax_game}
    \vthetat^*, \vphi^*, \lambdaco^* \triangleq    \underset{\vthetat, \vphi}{\text{argmin}}\ \underset{\lambdaco \ge 0}{\text{argmax}} \, \, \mathcal{L}_{\mathbf{I}}(\vthetat \odot \hat{\vz}(\vphi))  + \lambdaco \left( \texttt{BPP}(\texttt{cast}(\vthetat \odot \hat{\vz}(\vphi)) - \tau_{\texttt{BPP}} \right).
\end{equation}
We apply \textit{simultaneous} gradient descent
on ($\vthetat,\vphi$) and projected (to $\mathbb{R}^+)$ gradient ascent on $\lambdaco$. Note that unlike the penalized formulation in which the multiplicative factor for the constraint is fixed, here it is dynamically adjusted throughout the optimization. 
As detailed in \cref{app:proxy_constraints}, the medians face differentiability issues when they saturate at 0 or 1. Thus we employ the \textit{expected}  $\texttt{BPP}(\vthetat \odot \bar{\vz}(\vphi))$ as a proxy-constraint \citep{cotter2019} for computing the gradient for the update of the primal parameters.

Our constrained formulation grants a direct control over the compression rate of the final model. This provides an algorithmic approach to retain the best possible performance \textit{and} achieve a specific target BPP, without having to perform searches over the hyper-parameter $\lambdapen$ or the model architecture.

\section{Experiments}
\label{sec:experiments}

We perform experiments on the Kodak image dataset \citep{kodak} consisting of 24 images of size 768 $\times$ 512. We compare our approach with various image compression techniques: COIN \citep{dupont2021coin}, unstructured magnitude pruning with fine-tuning and JPEG \citep{wallace1992jpeg}. Details on the implementation and hyper-parameter configurations can be found in \cref{app:exp_details}. Comprehensive experiments and qualitative comparisons are presented in \cref{app:results}.

We make the difficulty of the task equal across techniques. For example, for a BPP budget of 0.3, we train a COIN model of dimensions 2-10x[28]-3. Thus, COIN models are fully dense and have a fixed BPP throughout training. Since L$_0$onie and magnitude pruning remove some of the parameters, we initialize them from larger models and require that they deliver a final model with a BPP of 0.3. The experimental settings used for each budget are provided in  \cref{app:exp_details}.

\begin{figure}[h]
    \hspace{-6mm}
    \begin{tabular}{ccc}
     \centering
     \hspace{3mm} {\small Image 8} & \hspace{3mm} {\small Image 14} & \hspace{3mm} {\small Image 22} \\
    \includegraphics[width=0.33\textwidth,trim={7mm 13mm 7mm 2mm},clip]{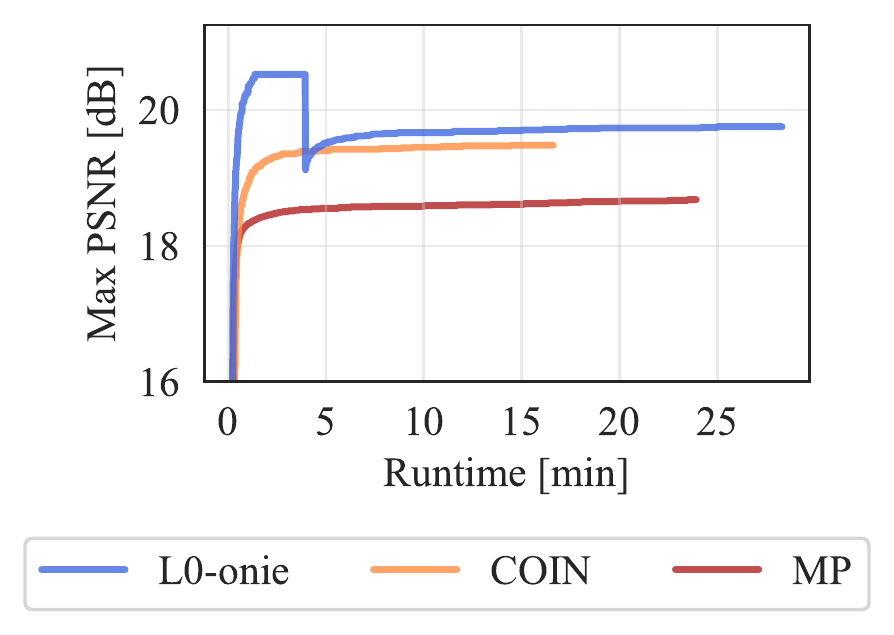} & \includegraphics[width=0.33\textwidth,trim={7mm 13mm 7mm 2mm},clip]{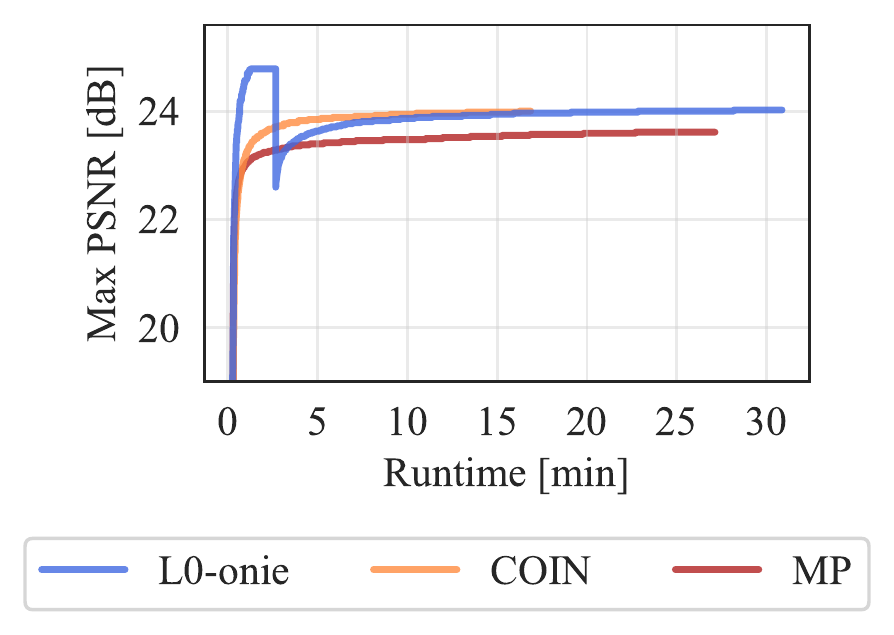} & \includegraphics[width=0.33\textwidth,trim={7mm 13mm 7mm 2mm},clip]{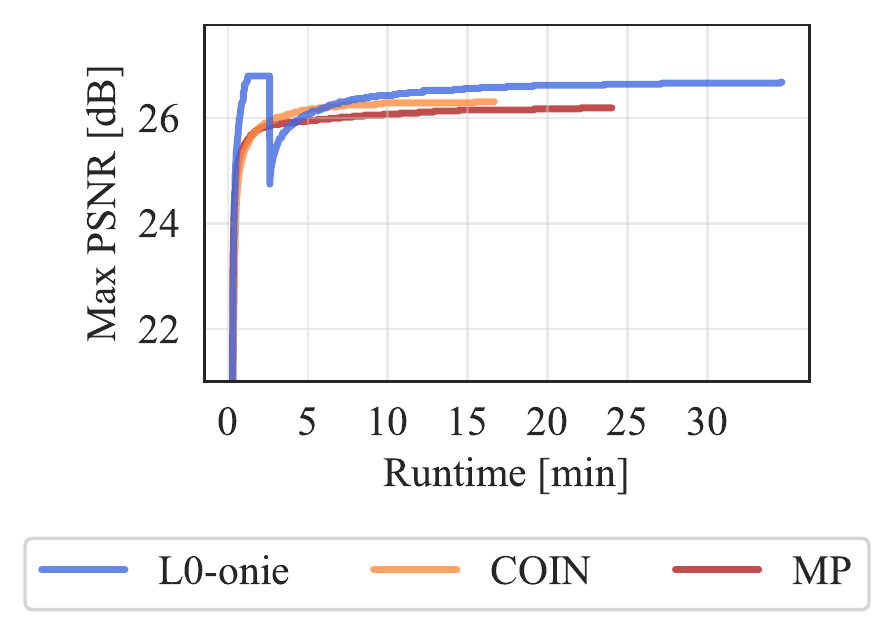}
    \end{tabular}
    \vspace{-2mm}
    \caption{Performance comparison during runtime for \loonie, \coin~and \mprune~with a target of 0.3 BPP. All methods are trained for 50k steps. Whenever the \loonie~model achieves the desired BPP, we reset its best PSNR meter. 
    }
    \label{fig:runtime_comparison}
\end{figure}

\textbf{Benchmark performance.} \cref{fig:psnr_vs_bpp} displays the performance of all methods at various BPP budgets. 
Magnitude pruning consistently under-performs all other approaches, both immediately after pruning (see \cref{fig:runtime_comparison}) and after fine-tuning. This is remarkable considering that we initialize magnitude pruning from a larger, fully trained COIN model.
L$_0$onie slightly surpasses the performance obtained by COIN.
On the other hand, although JPEG struggles at high compression rates it clearly dominates at larger BPPs; and its compression time is negligible in comparison to the other techniques.

\textbf{Higher performance, faster.} \cref{fig:runtime_comparison} shows the PSNR of the compressed model as a function of the training \textit{time}. As expected, the larger L$_0$onie model achieves better reconstructions much faster. However, recall that the original L$_0$onie model has a worse compression rate. To ensure a fair comparison, we refresh the PSNR meter for the L$_0$onie experiments as soon as they satisfy the BPP constraint -- this is marked by a drop in the maximum PSNR metric.

\vspace{-1ex}

Once the L$_0$onie models become feasible, they rapidly match and even surpass the performance of COIN. Despite each individual gradient step taking more time for the large L$_0$onie models, the L$_0$onie approach reaches a \textit{higher PSNR, faster than COIN}. Note that for the same wall-time budget as COIN, L$_0$onie achieves comparable or higher PSNR, while respecting the BPP constraint.

\begin{figure}[h]
    \centering
    \includegraphics[width=1.0\textwidth,trim={0mm 14mm 0mm 2mm},clip]{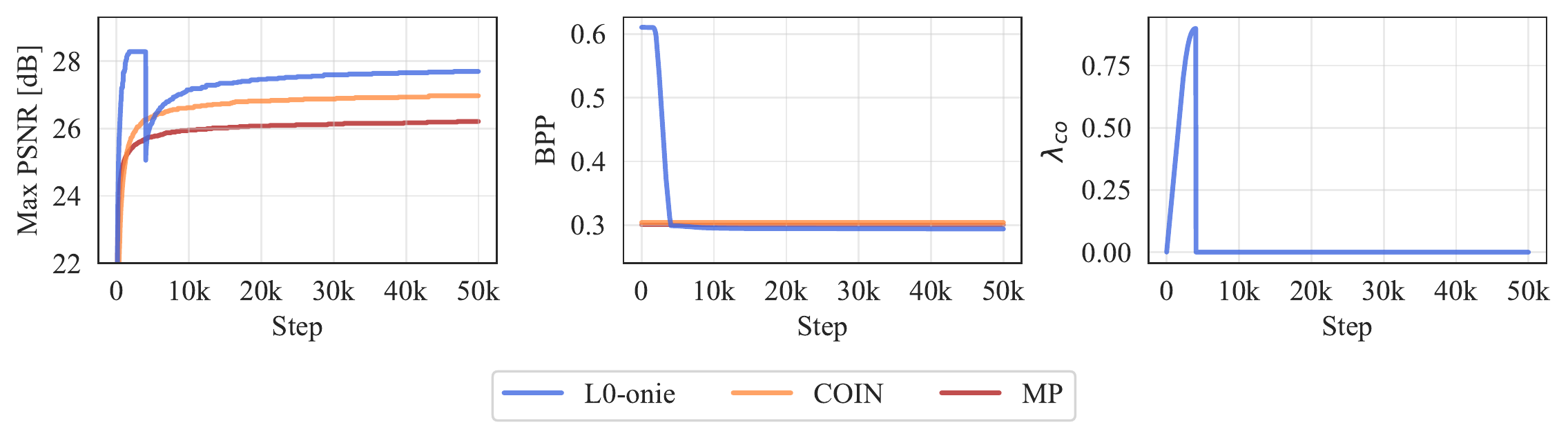}
    \vspace{-5mm}
    \caption{Training dynamics for \loonie, \coin~and \mprune~on image 7 of the Kodak dataset with a target BPP of 0.3.}
    \label{fig:dynamics_img15}
\end{figure}

\textbf{Training dynamics.}  \cref{fig:dynamics_img15} illustrates the training dynamics of all the methods. We include COIN and magnitude pruning experiments as baselines, whose BPP is fixed during training. 
These techniques exhibit a similar behavior, with fast initial growth, followed by a slow saturation period. 

L$_0$onie is initialized from a model with an unfeasible BPP. This causes the Lagrange multiplier $\lambda_{\text{co}}$ to grow at the beginning of training. In turn, the growing multiplier exerts a force that leads the optimization to trade-off reconstruction performance in order to reduce the model's BPP. Note that this feasibility is obtained relatively early during training. After the initial drop in PSNR, the now feasible model can focus on optimizing the reconstruction performance. We observe that once feasible, the PSNR for L$_0$onie model quickly recovers and surpasses that of COIN.

Note that after reaching feasibility, the BPP statistics barely change for the L$_0$onie model. Thus optimization is effectively taking place over a particular subnetwork. Pruning the network at this stage and ``fusing'' the gates and weights could allow for more efficient training. These gains could be particularly substantial when using structured sparsity, which was beyond the scope of our work.

\section{Limitations and Future work}

Recent work has improved the efficiency of COIN by considering a meta-learning framework \citep{dupont2022coinpp, lee2021meta}. As demonstrated by \citet{schwarz2022meta}, sparsity and meta-learning are complementary avenues for improving INR-based compression. Despite the controllability advantages of our constrained approach, further research is required for understanding the optimization dynamics of INRs, both in the constrained and unconstrained settings. 

For simplicity we concentrated on the unstructured sparsity case in this work. However, the compression gains resulting from structured sparsity (for example, grouping all weights associated with an input neuron under a shared gate) could be more notorious and more easily materialized. This is particularly important during training: once the INR becomes sparse, its unused parameters can be removed to enable faster training.

Finally, our sparsity-constrained formulation can support progressive sparsification of the model to reach different compression rates. Once a model has achieved a desired BPP, this very same model can be used in a straightforward manner as a starting point to obtain INRs with higher compression rates. It is not immediately evident how to achieve this progressive sparsification using vanilla COIN models or penalized L$_0$ formulations, while reliably controling the resulting compression rate.

\section{Conclusion}

We propose L$_0$onie, a sparse extension of COIN models trained using a constrained formulation. Our method allows to leverage the faster learning of overparameterized networks, while respecting a desired compression rate, without requiring costly hyper-parameter tuning or architecture search.

\section*{Acknowledgements}
JGP was funded by the Canada CIFAR AI Chair Program and by an IVADO Excellence Scholarship. 

\cref{fig:loonie_comparison} depicts a circulating unit of currency. Its use in this work contends as fair use for it is limited to a commentary relating to the \textit{image} of the currency itself. The image of all Canadian coins remains the copyright of the Royal Canadian Mint.

\bibliography{references.bib, loonie_references.bib}

\begin{thebibliography}{26}
\providecommand{\natexlab}[1]{#1}
\providecommand{\url}[1]{\texttt{#1}}
\expandafter\ifx\csname urlstyle\endcsname\relax
  \providecommand{\doi}[1]{doi: #1}\else
  \providecommand{\doi}{doi: \begingroup \urlstyle{rm}\Url}\fi

\bibitem[Arora et~al.(2018)Arora, Cohen, and
  Hazan]{aroraOptimizationDeepNetworks2018a}
S.~Arora, N.~Cohen, and E.~Hazan.
\newblock On the {{Optimization}} of {{Deep Networks}}: {{Implicit
  Acceleration}} by {{Overparameterization}}.
\newblock In \emph{{{ICML}}}, 2018.

\bibitem[Boyd and Vandenberghe(2004)]{boyd2004convex}
S.~Boyd and L.~Vandenberghe.
\newblock \emph{{Convex Optimization}}.
\newblock Cambridge University Press, 2004.

\bibitem[Bégaint et~al.(2020)Bégaint, Racapé, Feltman, and
  Pushparaja]{compressai}
J.~Bégaint, F.~Racapé, S.~Feltman, and A.~Pushparaja.
\newblock {CompressAI: a PyTorch library and evaluation platform for end-to-end
  compression research}.
\newblock \emph{arXiv:2201.12904}, 2020.

\bibitem[Cotter et~al.(2019)Cotter, Jiang, Gupta, Wang, Narayan, You, and
  Sridharan]{cotter2019}
A.~Cotter, H.~Jiang, M.~Gupta, S.~Wang, T.~Narayan, S.~You, and K.~Sridharan.
\newblock {Optimization with Non-Differentiable Constraints with Applications
  to Fairness, Recall, Churn, and Other Goals}.
\newblock In \emph{{JMLR}}, 2019.

\bibitem[Dupont et~al.(2021)Dupont, Golinski, Alizadeh, Teh, and
  Doucet]{dupont2021coin}
E.~Dupont, A.~Golinski, M.~Alizadeh, Y.~W. Teh, and A.~Doucet.
\newblock {{COIN}}: {{COmpression}} with {{Implicit Neural}} representations.
\newblock In \emph{{{ICLR}} - {{Neural Compression Workshop}}}, 2021.

\bibitem[Dupont et~al.(2022{\natexlab{a}})Dupont, Loya, Alizadeh, Goli{\'n}ski,
  Teh, and Doucet]{dupont2022coinpp}
E.~Dupont, H.~Loya, M.~Alizadeh, A.~Goli{\'n}ski, Y.~W. Teh, and A.~Doucet.
\newblock {{COIN}}++: {{Data Agnostic Neural Compression}}.
\newblock \emph{arXiv:2201.12904}, 2022{\natexlab{a}}.

\bibitem[Dupont et~al.(2022{\natexlab{b}})Dupont, Teh, and
  Doucet]{dupont2021generative}
E.~Dupont, Y.~W. Teh, and A.~Doucet.
\newblock {Generative Models as Distributions of Functions}.
\newblock In \emph{{AISTATS}}, 2022{\natexlab{b}}.

\bibitem[Gale et~al.(2019)Gale, Elsen, and Hooker]{gale2019state}
T.~Gale, E.~Elsen, and S.~Hooker.
\newblock {The State of Sparsity in Deep Neural Networks}.
\newblock In \emph{{NeurIPS - Workshop in ODML-CDNNR}}, 2019.

\bibitem[Gallego-Posada and Ramirez(2022)]{cooper}
J.~Gallego-Posada and J.~Ramirez.
\newblock {Cooper: a toolkit for Lagrangian-based constrained optimization}.
\newblock \texttt{\url{https://github.com/cooper-org/cooper}}, 2022.

\bibitem[Gallego-Posada et~al.(2021)Gallego-Posada, Ramirez, and
  Erraqabi]{gallego2021flexible}
J.~Gallego-Posada, J.~Ramirez, and A.~Erraqabi.
\newblock {Flexible Learning of Sparse Neural Networks via Constrained $L_0$
  Regularization}.
\newblock In \emph{{NeurIPS - LatinX in AI Workshop}}, 2021.

\bibitem[Jang et~al.(2017)Jang, Gu, and Poole]{jang2016categorical}
E.~Jang, S.~Gu, and B.~Poole.
\newblock {Categorical Reparameterization with Gumbel-Softmax}.
\newblock In \emph{{ICLR}}, 2017.

\bibitem[Kingma and Ba(2015)]{adam}
D.~P. Kingma and J.~Ba.
\newblock {Adam: A Method for Stochastic Optimization}.
\newblock In \emph{{ICLR}}, 2015.

\bibitem[Kodak(1991)]{kodak}
Kodak.
\newblock Kodak dataset.
\newblock \url{http://r0k.us/graphics/kodak/}, 1991.

\bibitem[Lee et~al.(2021)Lee, Tack, Lee, and Shin]{lee2021meta}
J.~Lee, J.~Tack, N.~Lee, and J.~Shin.
\newblock {Meta-Learning Sparse Implicit Neural Representations}.
\newblock In \emph{{NeurIPS}}, 2021.

\bibitem[Li et~al.(2017)Li, Kadav, Durdanovic, Samet, and
  Graf]{li2017l1pruning}
H.~Li, A.~Kadav, I.~Durdanovic, H.~Samet, and H.~P. Graf.
\newblock {Pruning Filters for Efficient ConvNets}.
\newblock In \emph{{ICLR}}, 2017.

\bibitem[Li et~al.(2021)Li, Niklaus, Snavely, and Wang]{li2021neural}
Z.~Li, S.~Niklaus, N.~Snavely, and O.~Wang.
\newblock {Neural Scene Flow Fields for Space-Time View Synthesis of Dynamic
  Scenes}.
\newblock In \emph{CVPR}, 2021.

\bibitem[Louizos et~al.(2018)Louizos, Welling, and Kingma]{louizos2017learning}
C.~Louizos, M.~Welling, and D.~P. Kingma.
\newblock {Learning Sparse Neural Networks through $L_0$ Regularization}.
\newblock In \emph{{ICLR}}, 2018.

\bibitem[Maddison et~al.(2017)Maddison, Mnih, and Teh]{maddison2016concrete}
C.~J. Maddison, A.~Mnih, and Y.~W. Teh.
\newblock {The Concrete Distribution: A Continuous Relaxation of Discrete
  Random Variables}.
\newblock In \emph{{ICLR}}, 2017.

\bibitem[Mescheder et~al.(2019)Mescheder, Oechsle, Niemeyer, Nowozin, and
  Geiger]{mescheder2019occupancy}
L.~Mescheder, M.~Oechsle, M.~Niemeyer, S.~Nowozin, and A.~Geiger.
\newblock {Occupancy Networks: Learning 3d Reconstruction in Function Space}.
\newblock In \emph{CVPR}, 2019.

\bibitem[Paszke et~al.(2019)Paszke, Gross, Massa, Lerer, Bradbury, Chanan,
  Killeen, Lin, Gimelshein, Antiga, Desmaison, Kopf, Yang, DeVito, Raison,
  Tejani, Chilamkurthy, Steiner, Fang, Bai, and Chintala]{pytorch}
A.~Paszke, S.~Gross, F.~Massa, A.~Lerer, J.~Bradbury, G.~Chanan, T.~Killeen,
  Z.~Lin, N.~Gimelshein, L.~Antiga, A.~Desmaison, A.~Kopf, E.~Yang, Z.~DeVito,
  M.~Raison, A.~Tejani, S.~Chilamkurthy, B.~Steiner, L.~Fang, J.~Bai, and
  S.~Chintala.
\newblock {PyTorch: An Imperative Style, High-Performance Deep Learning
  Library}, 2019.

\bibitem[Perez et~al.(2017)Perez, Strub, de~Vries, Dumoulin, and
  Courville]{perez2017film}
E.~Perez, F.~Strub, H.~de~Vries, V.~Dumoulin, and A.~Courville.
\newblock {FiLM: Visual Reasoning with a General Conditioning Layer}.
\newblock In \emph{{AAAI}}, 2017.

\bibitem[Schwarz and Teh(2022)]{schwarz2022meta}
J.~R. Schwarz and Y.~W. Teh.
\newblock {Meta-Learning Sparse Compression Networks}.
\newblock \emph{arXiv:2205.08957}, 2022.

\bibitem[Sitzmann et~al.(2020)Sitzmann, Martel, Bergman, Lindell, and
  Wetzstein]{sitzmann2020implicit}
V.~Sitzmann, J.~Martel, A.~Bergman, D.~Lindell, and G.~Wetzstein.
\newblock {Implicit Neural Representations with Periodic Activation Functions}.
\newblock In \emph{{NeurIPS}}, 2020.

\bibitem[Stanley(2007)]{stanley2007compositional}
K.~O. Stanley.
\newblock {Compositional Pattern Producing Networks: A Novel Abstraction of
  Development}.
\newblock \emph{Genetic Programming and Evolvable Machines}, 8\penalty0
  (2):\penalty0 131--162, 2007.

\bibitem[Tancik et~al.(2020)Tancik, Srinivasan, Mildenhall, {Fridovich-Keil},
  Raghavan, Singhal, Ramamoorthi, Barron, and Ng]{tancik2020FourierFeatures}
M.~Tancik, P.~P. Srinivasan, B.~Mildenhall, S.~{Fridovich-Keil}, N.~Raghavan,
  U.~Singhal, R.~Ramamoorthi, J.~T. Barron, and R.~Ng.
\newblock Fourier {{Features Let Networks Learn High Frequency Functions}} in
  {{Low Dimensional Domains}}.
\newblock In \emph{{NeurIPS}}, 2020.

\bibitem[Wallace(1992)]{wallace1992jpeg}
G.~Wallace.
\newblock {The JPEG still picture compression standard}.
\newblock \emph{IEEE Transactions on Consumer Electronics}, 38\penalty0
  (1):\penalty0 xviii--xxxiv, 1992.

\end{thebibliography}
\bibliographystyle{support_files/abbrvnatClean}

\newpage

\appendix

\section{Stochastic Gates}
\label{app:gates}

Let $U_j \sim \text{Unif}(0, 1)$ and $0 < \beta < 1$. A concrete random variable \citep{maddison2016concrete, jang2016categorical} $s_j \sim q(\cdot\ |\ (\phi_j, \beta))$ can be obtained by transforming the uniform random variable as:
\begin{equation}
    \label{eq:concrete}
    s_j = \text{Sigmoid} \left( \frac{1}{\beta} \log \left( \frac{\phi_j \, U_j}{1-U_j} \right) \right).
\end{equation}

Given hyper-parameters $\gamma < 0 < 1 < \zeta$, the hard concrete distribution \citep{louizos2017learning} corresponds to a stretching and clamping of a concrete random variable. The hard concrete distribution is a mixed distribution with point masses at $0$ and $1$, and a continuous density over $(0, 1)$.
\begin{equation}
    \label{eq:hard_concrete}
     \hspace{5mm}
     \vz = \text{clamp}_{[0, 1]}( \vect{s} (\zeta - \gamma) + \gamma))
 \end{equation}

\cref{tab:gates_params} specifies the values of the fixed parameters associated with the gates employed throughout this work, following \citet{louizos2017learning}. 

\begin{table}[h]
    \renewcommand{\arraystretch}{1.1}
    \centering
    \caption{Fixed parameters of the hard concrete distribution.}
    \label{tab:gates_params}
    \begin{tabular}{|c|ccc|}
      \hline
      \textbf{Parameter} & $\gamma$ & $\zeta$ & $\beta$ \\
      \hline
      \textbf{Value} & -0.1 & 1.1 & 2/3 \\
      \hline
    \end{tabular}
\end{table}

We use $\vz$ for modeling the stochastic gates of L$_0$onie models. The stochastic nature of hard concrete variables entails a model which is itself \emph{stochastic}. For computing forwards and BPP measurements, we replace each gate by its median $\hat{\vz}(\vphi)$: 
\begin{equation}
    \label{eq:medians}
    \hat{\vz}(\vphi) = \text{min}\left(1,\text{ max}\left(0,\ \text{Sigmoid}\left(\frac{\text{log}(\vphi)} {\beta}\right) (\zeta - \gamma) + \gamma \right)\right)
\end{equation}

Gate medians are a deterministic function of the trainable parameter $\vphi$. Moreover, the stretching and clamping enables the medians to attain the values 0 or 1, thus producing a sparse network. 

However, gate medians are poorly suited for setting up the BPP constraint as they do not allow for change once a gate is ``fully turned off or on''; once the median is zero, the gradient with respect to $\phi_j$ is zero. Hence, we use expected BPP of the model as a surrogate for computing gradients. Note that this quantity is differentiable with respect to $\phi_j$.
\begin{equation*}
    \label{eq:exp_bpp}
    \mathbb{E}_{\vz|\bg{\phi}}\left[\ \texttt{BPP}(\vthetat \odot \vz) \right] = \sum_{i} \texttt{bits}(\vthetat_i \odot \vz_j)\ \mathbb{P}[ z_i \neq 0 ] = \sum_{i} \texttt{bits}(\vthetat_i) \, \text{Sigmoid}\left( \log (\phi_i) - \beta \log \frac{- \gamma}{\zeta} \right)
\end{equation*}

\section{Proxy-constraints}
\label{app:proxy_constraints}

Let us recall the min-max Lagrangian optimization problem from \eqref{eq:minmax_game}
\begin{equation*}
    \vthetat^*, \vphi^*, \lambdaco^* \triangleq    \underset{\vthetat, \vphi}{\text{argmin}}\ \underset{\lambdaco \ge 0}{\text{argmax}} \, \, \mathcal{L}_{\mathbf{I}}(\vthetat \odot \hat{\vz}(\vphi))  + \lambdaco \left( \texttt{BPP}(\texttt{cast}(\vthetat \odot \hat{\vz}(\vphi)) - \tau_{\texttt{BPP}} \right).
\end{equation*}

Applying simultaneous gradient descent-ascent on this problem corresponds to the update scheme:
\begin{align}
    \label{eq:gda_updates}
    [\vthetat^{t+1}, \vphi^{t+1}] &\triangleq [\vthetat^t, \vphi^t] \ {\color{red} -}\ \eta_{\text{primal}} \,  \nabla_{[\vthetat, \vphi]} \left[ \mathcal{L}_{\mathbf{I}}(\vthetat \odot \hat{\vz}(\vphi))  + \lambdaco \texttt{BPP}(\texttt{cast}(\vthetat \odot \hat{\vz}(\vphi)) \right] \nonumber \\
    \lambdaco^{t+1} &\triangleq \max \left( 0, \lambdaco^{t}\ {\color{red} +} \ \eta_{\text{dual}} \,  \left( \texttt{BPP}(\texttt{cast}(\vthetat \odot \hat{\vz}(\vphi)) - \tau_{\texttt{BPP}} \right) \right) \nonumber
\end{align}

 The gradient update for $\lambdaco$ matches the value of constraint violation. Whenever the constraint is not satisfied, the Lagrange multiplier increases. We employ the dual restarts technique of \citet{gallego2021flexible} when the constraint is satisfied.
 
 Note that primal update involves computing the gradient of the $\texttt{BPP}(\texttt{cast}(\vthetat \odot \hat{\vz}(\vphi))$ with respect to the parameters $\vthetat$ and $\vphi$. As mentioned in \cref{app:gates}, this quantity is not differentiable when the gate medians attain the values 0 or 1. To overcome this issue we employ the \textit{expected} BPP as a surrogate or \textit{proxy-constraint} \citep[\S 4.2]{cotter2019} when computing the gradient for the primal update. The original (non-proxy) constraint is used for updating the value of the Lagrange multiplier.

\section{Experimental Details}

\label{app:exp_details}

Our implementation is developed in Python 3.8, using Pytorch 1.11 \citep{pytorch} and the Cooper constrained optimization library \citep{cooper}. We provide scripts to replicate our experiments at: \texttt{\url{https://github.com/juan43ramirez/l0onie}}.

\subsection{Codec Baselines}
\label{app:exp_details:codec}

The baseline results for the JPEG \citep{wallace1992jpeg} standard were obtained using the  CompressAI library \citep{compressai}.

\subsection{Model Architectures}
\label{app:exp_details:model_archs}

We considered the same setup as \citet{dupont2021coin}: MLPs with 2 input dimensions corresponding to the position of input pixels normalized to lie in $[-1, 1]$ and 3 outputs corresponding to RGB values normalized to lie in $[0, 1]$. These models are trained at single (\texttt{float32}) precision.

All our models use Sine activations at every layer except for the output layer. These are setup with a fixed angular frequency of $\omega_0 = 30$. 

When replicating the experiments of \citet{dupont2021coin}, we consider specific architectures so as to yield bits-per-pixel of $0.07$, $0.15$, $0.3$ and $0.6$ at half (\texttt{float16}) precision. Table \ref{tab:base_architectures} contains the hidden layers associated with each of these architectures.

\begin{table}[h!]
    \centering
    \begin{tabular}{cccc}
        \textbf{Hidden layers} & \textbf{Width of Layers} & \textbf{BPP @ \texttt{float32}} & \textbf{BPP @ \texttt{\texttt{float16}}}  \\
        \hline
        \hline
        5 & 20 & 0.15 & 0.07 \\
        5 & 30 & 0.3 & 0.15  \\
        10 & 28 & 0.6 & 0.3  \\
        10 & 40 & 1.2 & 0.6  \\
        10 & 40 & 1.62 & 0.81 \\
        \hline
    \end{tabular}
    \caption{Architectures used throughout this paper, along with their respective bits-per-pixel when used to compress an image of 768 $\times$ 512 pixels. BPPs are calculated for each architecture at single (\texttt{float32}) and half (\texttt{\texttt{float16}}) precision.}
    \label{tab:base_architectures}
\end{table}

For L$_0$onie and magnitude pruning experiments, we consider \textit{unstructured} sparsity, where each parameter can be pruned individually. We also consider \textit{target BPPs} of compressed models at the same levels of $0.07$, $0.15$, $0.3$ and $0.6$ used to evaluate the COIN approach. Furthermore, the \textit{initial} (dense) architectures which are sparsified by L$_0$onie and magnitude pruning to meet these target BPPs are also based on the architectures presented in Table \ref{tab:base_architectures}. We consider the next larger architecture from the one whose BPP matches the one targeted. This is illustrated in Table \ref{tab:target_architectures}.

\begin{table}[h!]
    \vspace{-1ex}
    \centering
    \begin{tabular}{cccc}
        \multirow{2}{*}{\textbf{Target BPP (@ \texttt{float16})}} & \multicolumn{3}{c}{\textbf{Initial Architecture}} \\
        \cline{2-4}
        & BPP (@ \texttt{float16}) & Hidden Layers & Width of Layers  \\
        \hline
        \hline
        0.07 & 0.15 & 5 & 30 \\
        0.15 & 0.3 & 10 & 28  \\
        0.3 & 0.6 & 10 & 40  \\
        0.6 & 0.81 & 13 & 40 \\
        \hline
    \end{tabular}
    \caption{Initial architectures considered when applying L$_0$onie and magnitude pruning to meet various target BPPs. Note that the initial architecture has a \textit{larger} BPP than the desired target.}
    \label{tab:target_architectures}
\end{table}

\subsection{Magnitude Pruning}

All magnitude pruning experiments follow the procedure presented in this section. First, we train a COIN model at a certain (larger) BPP. This serves as a baseline which can then be pruned to achieve a lower BPP. The selection of the initial architecture depends on the provided target BPP. The details about this choice are presented in \cref{tab:target_architectures}. 

Thereafter, we loop over the layers of the baseline model, sorting their individual parameters based on magnitude. A suitable proportion of the parameters with the smaller magnitude is set to 0. We perform the sorting and pruning separately for weight matrices and biases. 

We tried two variants for maginitude pruning: \blobletter{1} applying the \textit{same} level of pruning for all the layers, or \blobletter{2} keeping the first and last layers fully dense. This last technique is commonly used in conjunction with magnitude pruning, and is particularly important in the setting of this work since the number of input dimensions is very low.

The results of method \blobletter{1} were significantly and consistently worse than those of method \blobletter{2}. The results reported in this paper correspond to method \blobletter{2}. 

Immediately after pruning we evaluate the performance of the model in terms of PSNR. We identified a very significant degradation in performance and thus perform fine-tuning on the remaining parameters using the same number of iterations as the original training for the COIN baseline. The optimization hyper-parameters employed during this stage are presented in \ref{app:exp_details:optimization}.

\subsection{Parameter Initialization}

We use the same method proposed by \citep{louizos2017learning} for initializing the learnable parameters $\vphi$. We leverage this scheme to ensure that the expected value of stochastic gates is centered around $0.5$ at initialization.

The network's weights and biases are initialized in accordance to their use of sine activation functions, following the procedure described in \citep{sitzmann2020implicit}. This involves setting them based on a uniform distribution in $[-a, a]$. As mentioned previously, for L$_0$onie experiments, the gate distributions are initialized to be symmetric around 0.5, thus shrinking the effective value of parameters to lie in $[-a/2, a/2]$. We counter this by adjusting the initialization of weights and biases of L$_0$onie models to lie in $[-2a, 2a]$, thus having effective values of parameters in the desired range.

\subsection{Optimization Hyper-Parameters}
\label{app:exp_details:optimization}

This section presents the optimization hyper-parameters used throughout our work. \cref{tab:optim_coin_mp} indicates those considered when training COIN models and for the fine-tuning of models after magnitude pruning. These are the same across different images and architectures. Both cases employ the same configuration for the optimizer: Adam \citep{adam} with $(\beta_1,\ \beta_2) = (0.9,\ 0.99)$, as is default in Pytorch. 
\begin{table}[h]
    \vspace{-1ex}
    \renewcommand{\arraystretch}{1.1}
    \centering
    \caption{Optimization hyper-parameters employed when training COIN baselines and when fine-tuning models which have been magnitude pruned.}
    \label{tab:optim_coin_mp}
    \vspace{1ex}
    \begin{tabular}{cccccc}
    \hline
      \textbf{Approach} & \textbf{Train Precision} & \textbf{Decode Precision} & \textbf{Steps}  &  \textbf{Optim.} & $\weightlr$ \\
      \hline
      \hline
      COIN & \multirow{2}{*}{\texttt{float32}} & \multirow{2}{*}{\texttt{float16}}  & \multirow{2}{*}{50.000} & \multirow{2}{*}{Adam} & \multirow{2}{*}{$2 \cdot 10^{-4}$}\\ 
      MP fine-tuning & &  & &  &  \\ 
       \hline
      
    \end{tabular}
\end{table}

Note that the constraint functions considered for L$_0$onies involve expectations but can be computed in closed-form based on the parameters of the gates, as presented in \cref{app:gates}. Therefore, the computation of constraint violations is deterministic. We employ gradient \textit{ascent} on the Lagrange multipliers. We initialize all Lagrange multipliers at zero. Details on the chosen dual learning rate, along with the use of dual restarts for L$_0$onie experiments are presented in \cref{tab:optim_loonie}. 

We use the same configurations across all Kodak images for a given target BPP. Since our goal is to ``overfit'' the model to the image, we do not use weight decay or other regularization techniques.

\begin{table}[h!]
    \vspace{-1ex}
    \renewcommand{\arraystretch}{1.1}
    \centering
    \caption{Optimization hyper-parameters employed when training L$_0$onie models.}
    \label{tab:optim_loonie}
    \vspace{1ex}
    \resizebox{\textwidth}{!}{%
    \begin{tabular}{ccccccccc}
    \hline
      \textbf{Target} & \textbf{Initial} & \multicolumn{2}{c}{\textbf{Weights}} & \multicolumn{2}{c}{\textbf{Gates}} & \multicolumn{3}{c}{\textbf{Lagrange Multipliers}} \\
      \cline{3-9} 
      \textbf{BPP} & \textbf{Architecture} & Optim. & $\weightlr$ & Optim. & $\gatelr$ & Optim. & $\duallr$ & Restarts \\ 
      \hline
      \hline
      0.07 & 2 - 5$\times$[30] - 3 & \multirow{4}{*}{Adam} & \multirow{4}{*}{$10^{-3}$} & \multirow{4}{*}{Adam} & \multirow{4}{*}{$7 \cdot 10^{-4}$} & \multirow{4}{*}{Grad. Ascent} & $7 \cdot 10^{-3}$ & \multirow{4}{*}{Yes} \\ 
      0.15 & 2 - 10$\times$[28] - 3 & & & & & & $3 \cdot 10^{-3}$ \\
      0.3 & 2 - 10$\times$[40] - 3 & & & & & & $1 \cdot 10^{-3}$ \\
      0.6 & 2 - 13$\times$[40] - 3 & & & & & & $8 \cdot 10^{-4}$ \\
       \hline
      
    \end{tabular}
    }
\end{table}

\newpage
\section{Additional Results}
\label{app:results}

\begin{figure}[h!]
    \begin{tabular}{cc}
         \rotatebox{90}{ \hspace{11mm} \textbf{0.07 BPP}} & \includegraphics[width=\textwidth,trim={0mm 10mm 0mm 2mm}, clip]{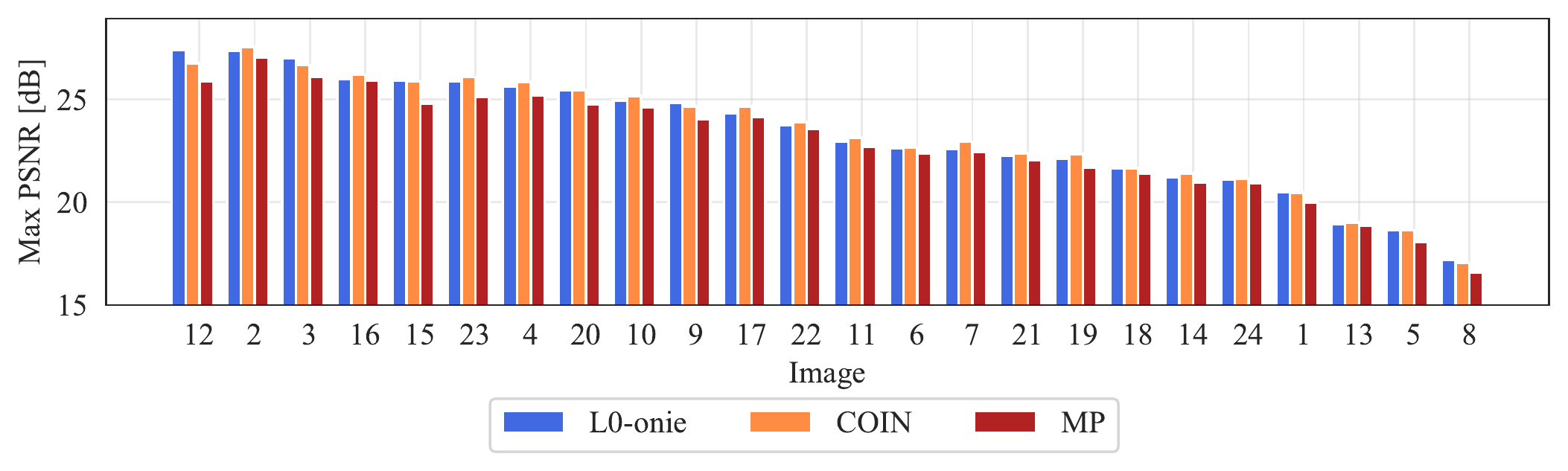} \\
         \rotatebox{90}{ \hspace{11mm} \textbf{0.15 BPP}} & \includegraphics[width=\textwidth,trim={0mm 10mm 0mm 2mm}, clip]{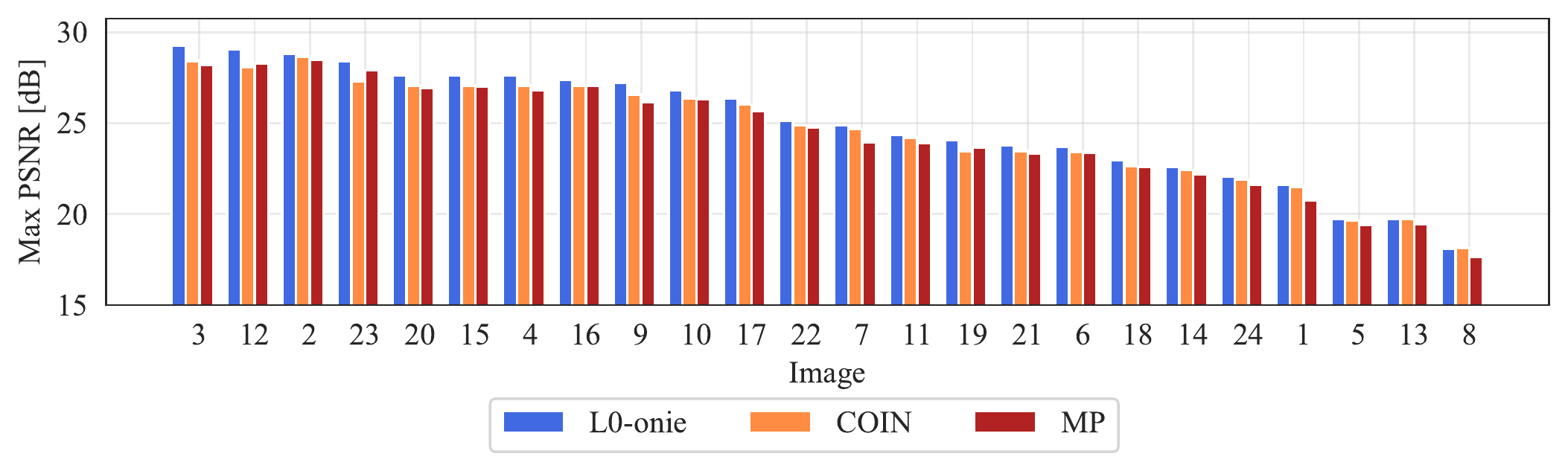} \\
         \rotatebox{90}{ \hspace{11mm} \textbf{0.3 BPP}} & \includegraphics[width=\textwidth,trim={0mm 10mm 0mm 2mm}, clip]{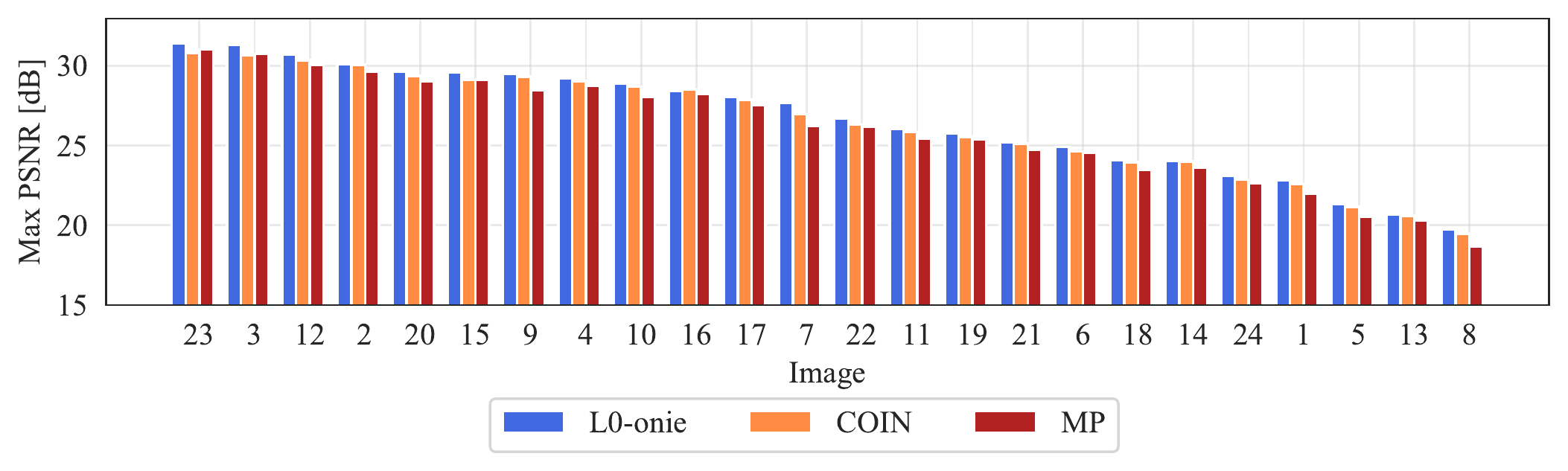} \\
         \rotatebox{90}{ \hspace{18mm} \textbf{0.6 BPP}} & \includegraphics[width=\textwidth,trim={0mm 2mm 0mm 2mm}, clip]{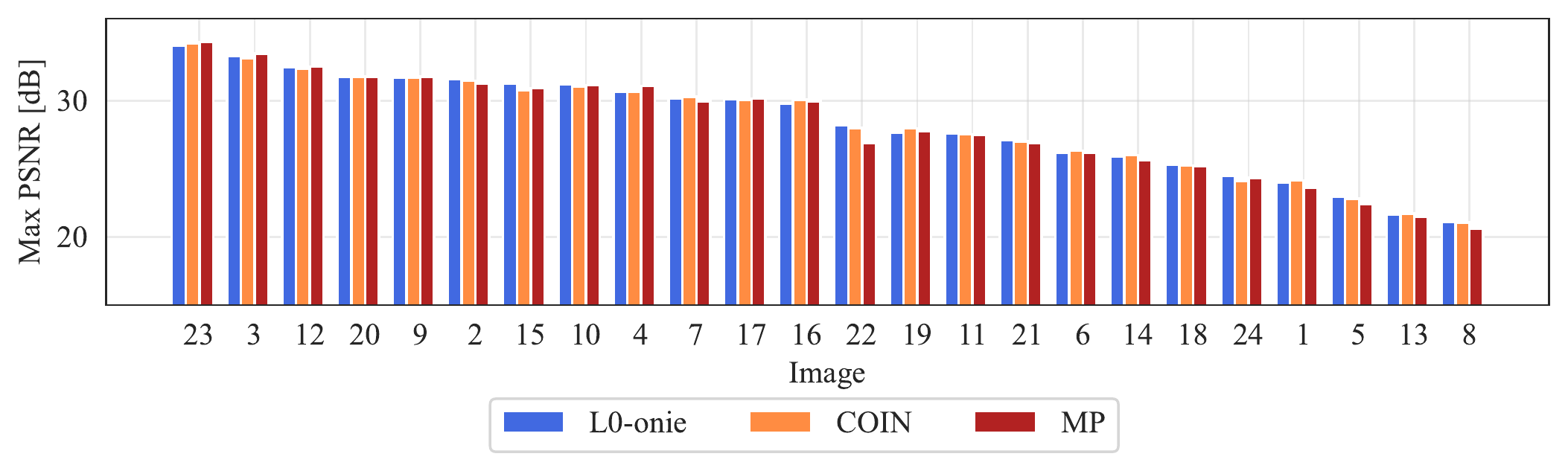}
    \end{tabular}
    \caption{Histogram of best PSNR for all images in the Kodak dataset for \loonie, \coin~and \mprune~at different compression rates. Images sorted in decreasing order of maximum PSNR for \loonie at the end of trainig.}
    \label{fig:hists}
\end{figure}

We evaluate the performance of our approach when compressing each image in the Kodak dataset at 0.07, 0.15, 0.3 and 0.6 BPPs. In addition, we did the same for COIN and magnitude pruning plus fine-tuning. \cref{fig:hists} presents the best PSNRs we obtained. 

Magnitude pruning always underperforms as opposed to L$_0$onie and COIN. This gap is more pronounced at 0.07 BPP, but is consistent across images. Moreover, the performance of L$_0$onie is generally competitive to that of COIN, whilst being slightly superior in the 0.15 and 0.3 BPP cases.

\newpage
\section{Qualitative Results}

We provide qualitative comparisons for some images of the Kodak dataset of different levels of compression difficulty, according to the histograms in \cref{fig:hists}. The image grids below are generated from reconstructions at various BPP budgets for \loonie, \coin, \mprune~and \jpeg. The experimental configuration for these experiments is provided in \cref{app:exp_details}.

\vspace{1.5cm}

\begin{figure}[h]
    \hspace{-28mm}
    \begin{tabular}{ccccc}
        \centering
        & & \multicolumn{2}{c}{ \rotatebox{90}{\hspace{6mm} \textbf{Original}} \includegraphics[width=0.3\textwidth,trim={7mm 12mm 7mm 2mm},clip]{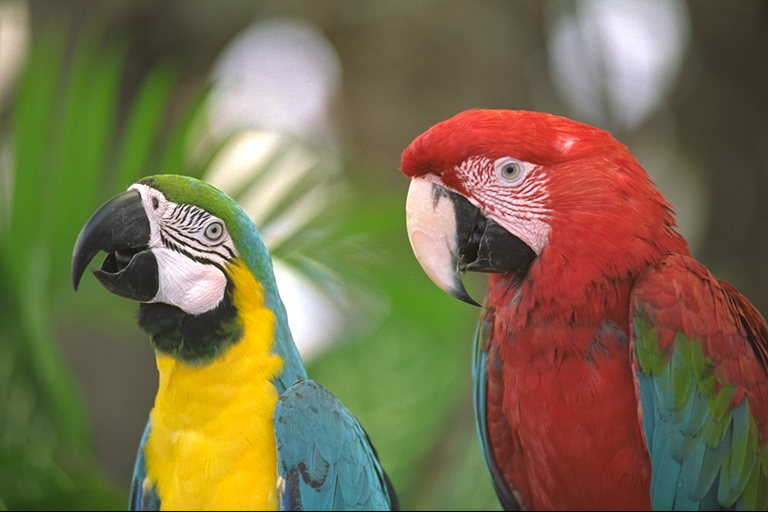}} & \\[3mm]
        \rotatebox{90}{ \hspace{7mm} \textbf{\loonie}} & \includegraphics[width=0.3\textwidth,trim={7mm 12mm 7mm 2mm},clip]{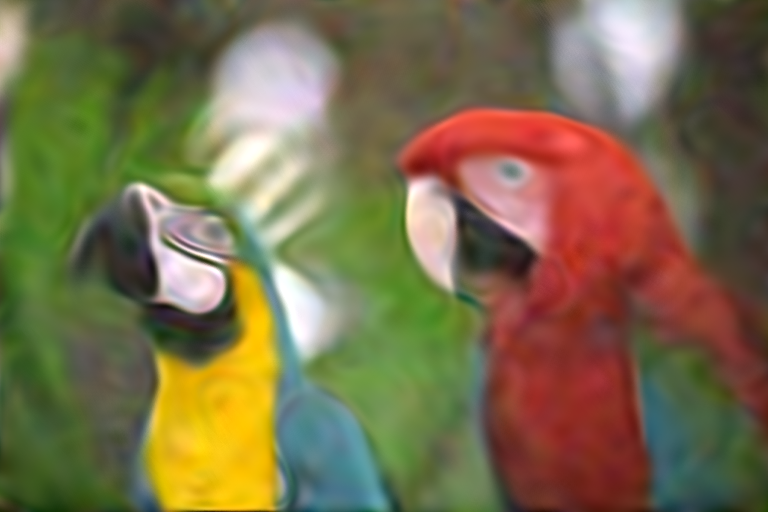} & \includegraphics[width=0.3\textwidth,trim={7mm 12mm 7mm 2mm},clip]{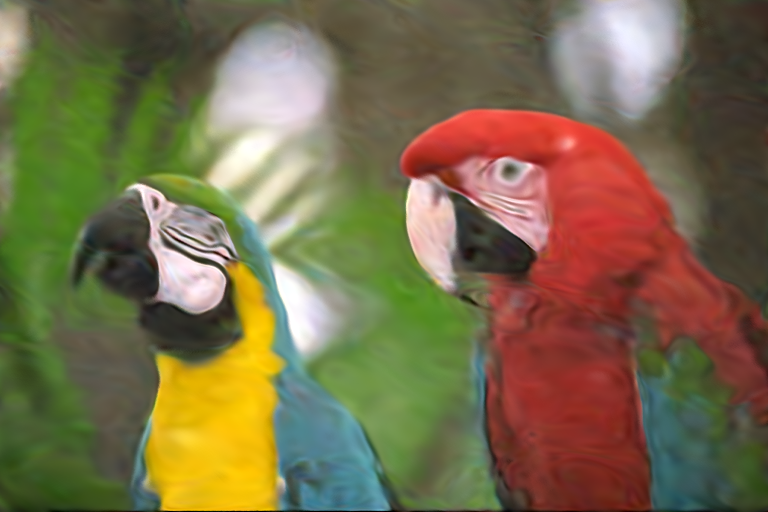} & \includegraphics[width=0.3\textwidth,trim={7mm 12mm 7mm 2mm},clip]{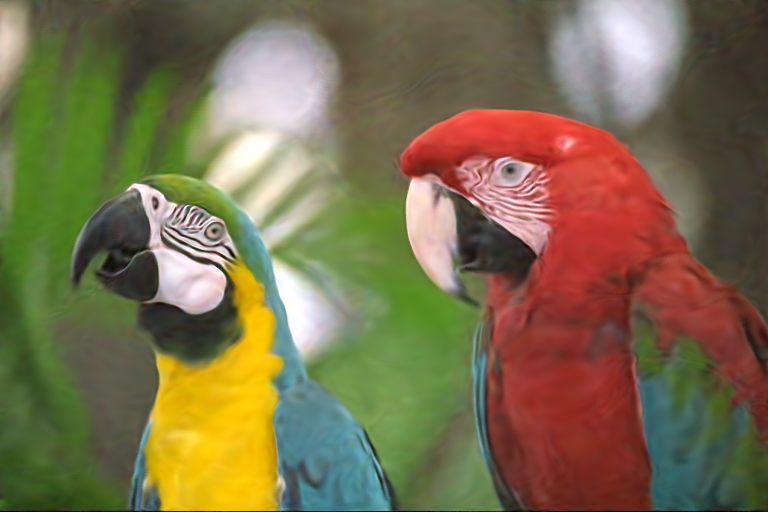} & \includegraphics[width=0.3\textwidth,trim={7mm 12mm 7mm 2mm},clip]{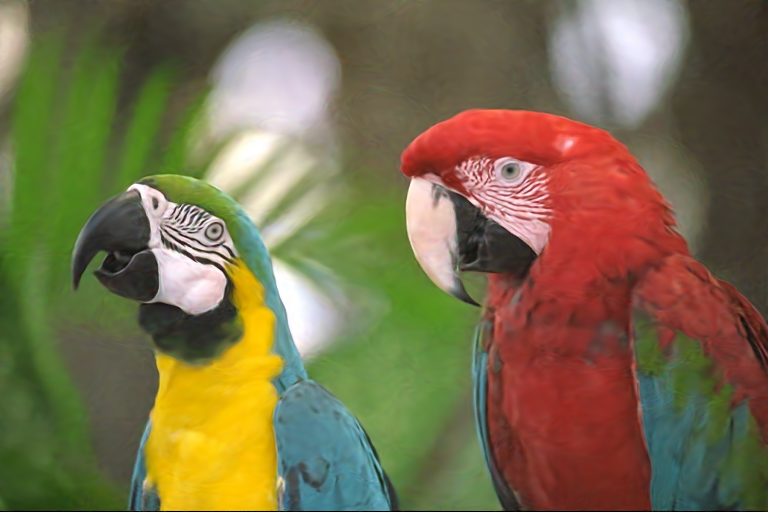} \\[3mm]
        \rotatebox{90}{ \hspace{8mm} \textbf{\coin}} & \includegraphics[width=0.3\textwidth,trim={7mm 12mm 7mm 2mm},clip]{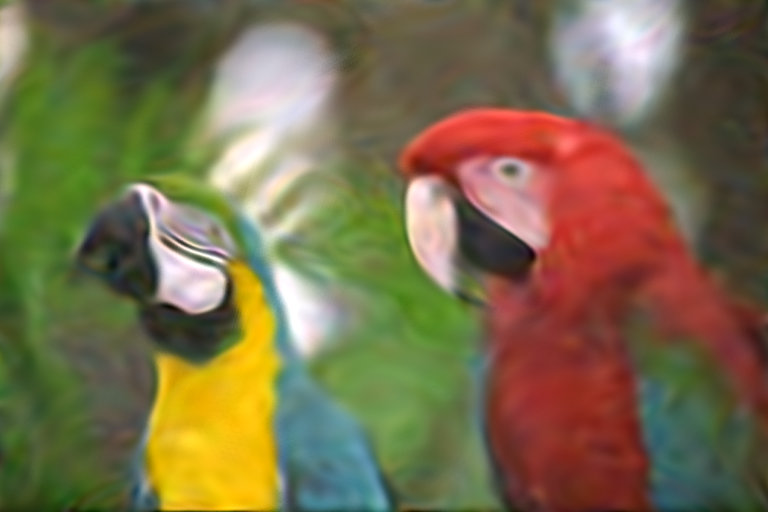} & \includegraphics[width=0.3\textwidth,trim={7mm 12mm 7mm 2mm},clip]{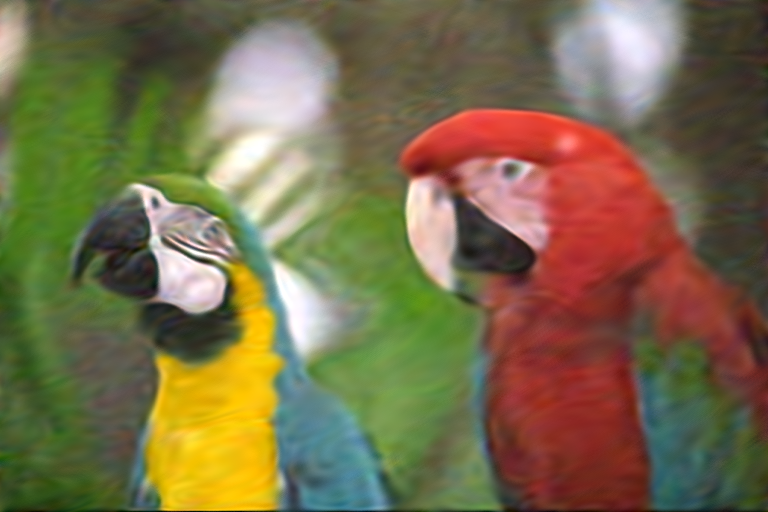} & \includegraphics[width=0.3\textwidth,trim={7mm 12mm 7mm 2mm},clip]{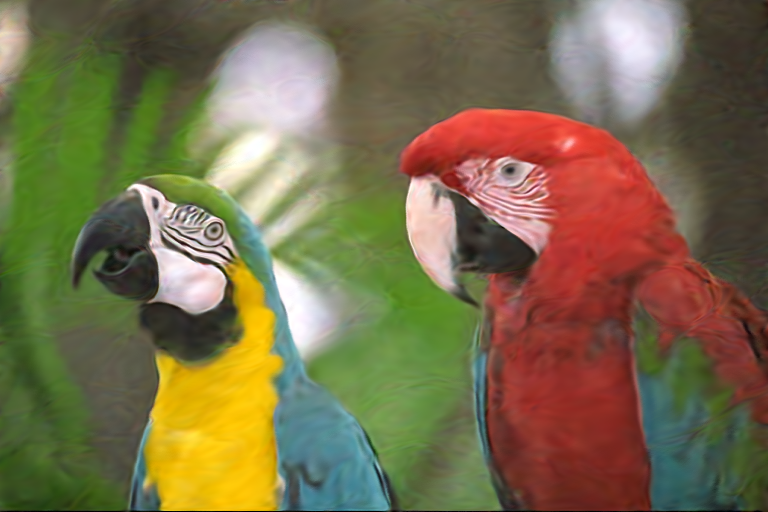} & \includegraphics[width=0.3\textwidth,trim={7mm 12mm 7mm 2mm},clip]{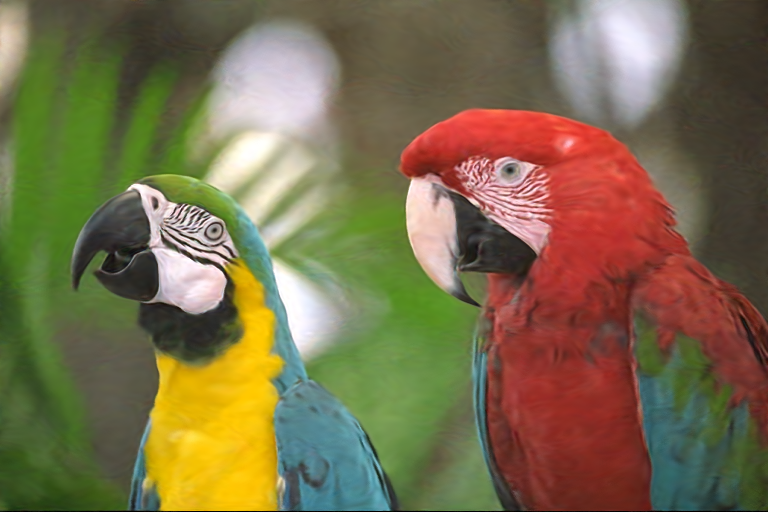} \\[3mm]
        \rotatebox{90}{ \hspace{9mm} \textbf{{\color{myred} MP}}} & \includegraphics[width=0.3\textwidth,trim={7mm 12mm 7mm 2mm},clip]{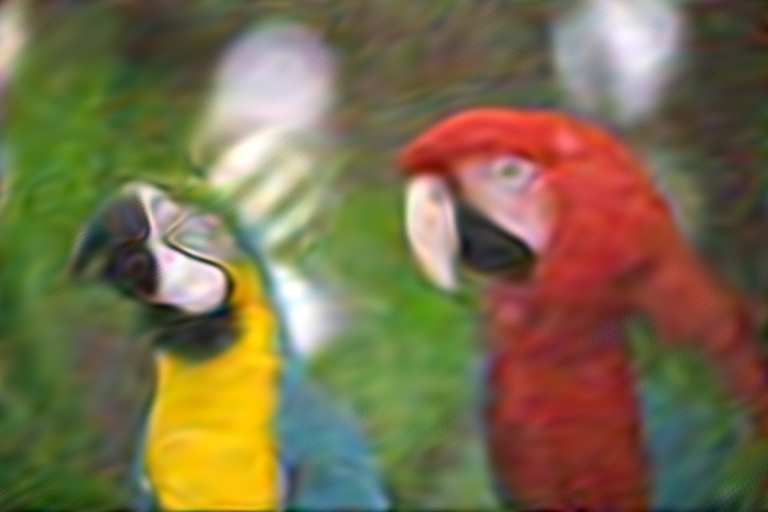} & \includegraphics[width=0.3\textwidth,trim={7mm 12mm 7mm 2mm},clip]{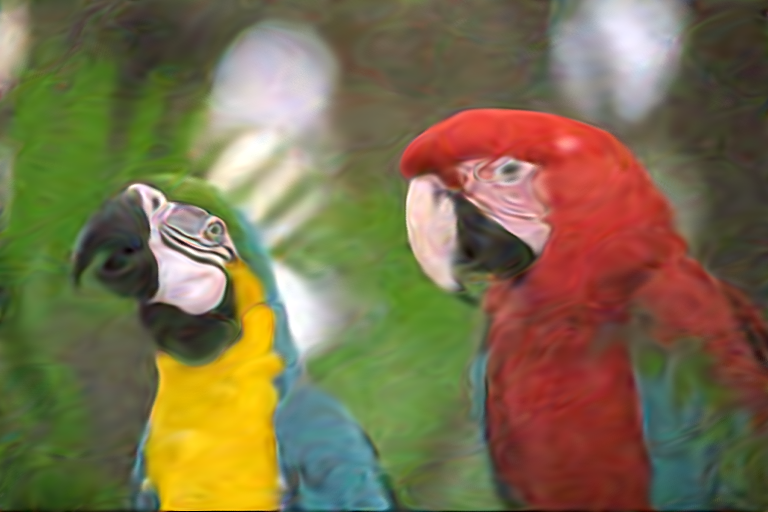} & \includegraphics[width=0.3\textwidth,trim={7mm 12mm 7mm 2mm},clip]{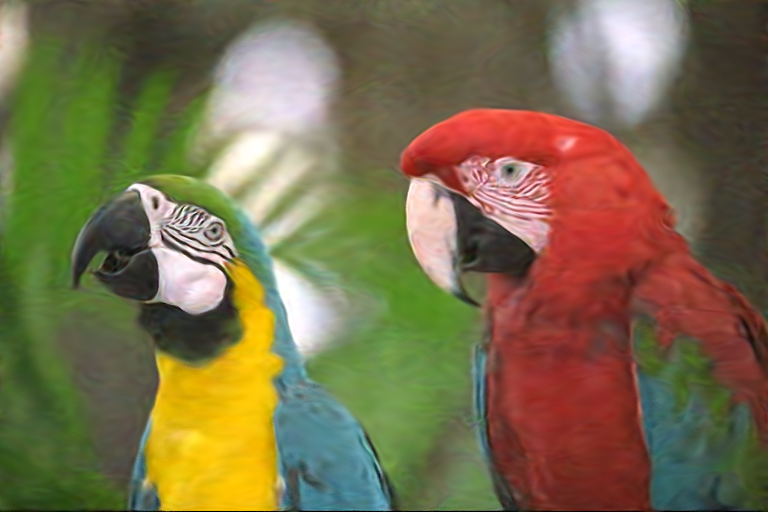} & \includegraphics[width=0.3\textwidth,trim={7mm 12mm 7mm 2mm},clip]{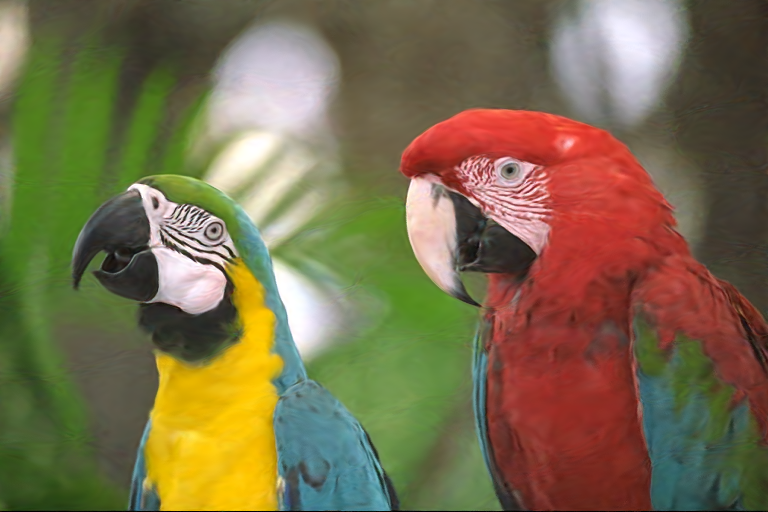} \\[3mm]
        \rotatebox{90}{ \hspace{8mm} \textbf{\jpeg}} & \includegraphics[width=0.3\textwidth,trim={7mm 12mm 7mm 2mm},clip]{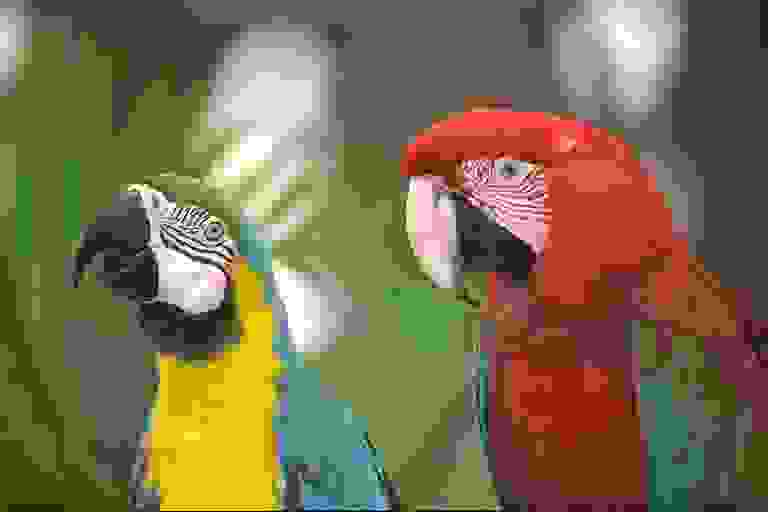} & \includegraphics[width=0.3\textwidth,trim={7mm 12mm 7mm 2mm},clip]{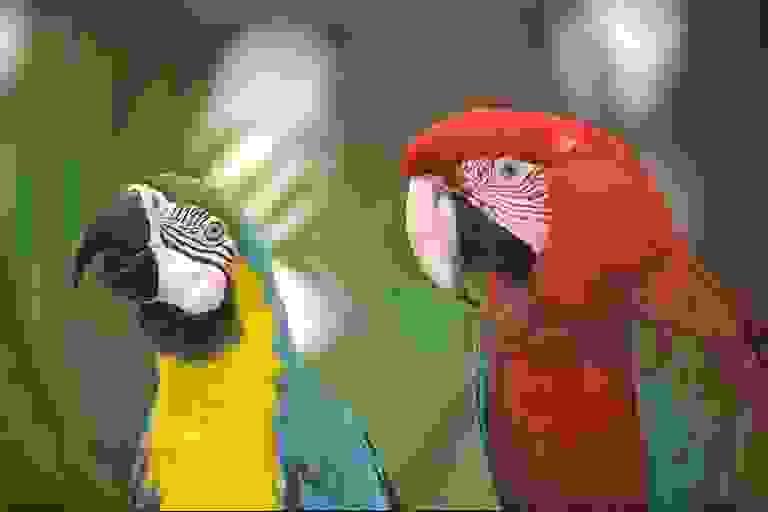} & \includegraphics[width=0.3\textwidth,trim={7mm 12mm 7mm 2mm},clip]{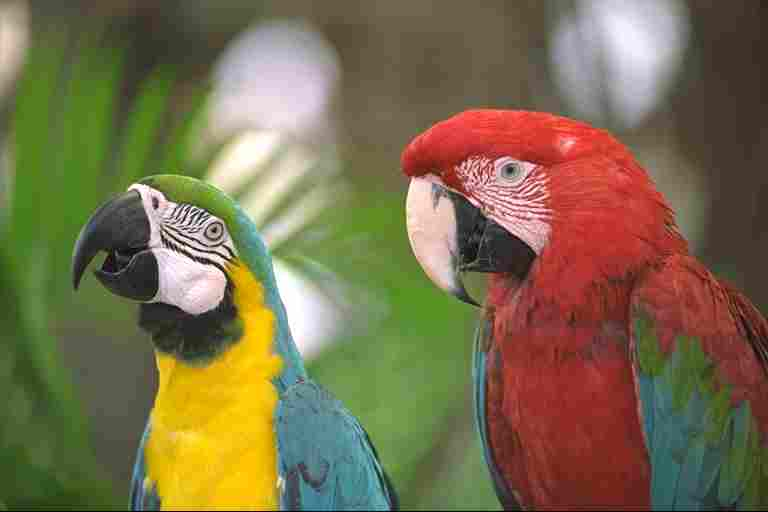} & \includegraphics[width=0.3\textwidth,trim={7mm 12mm 7mm 2mm},clip]{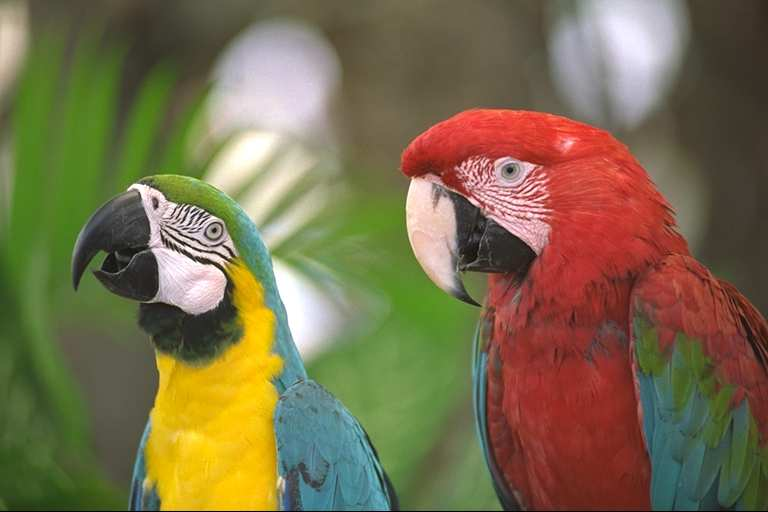} \\[3mm]
        & \textbf{0.07 BPP} & \textbf{0.15 BPP} & \textbf{0.3 BPP} & \textbf{0.6 BPP}
    \end{tabular}
    \vspace{-2mm}
    \caption{Reconstruction of Kodak image 23 for all methods at different target BPP.}
    \label{fig:kodim23}
\end{figure}

\begin{figure}[h]
    \hspace{-28mm}
    \begin{tabular}{ccccc}
        \centering
        & & \multicolumn{2}{c}{ \rotatebox{90}{\hspace{6mm} \textbf{Original}} \includegraphics[width=0.3\textwidth,trim={7mm 12mm 7mm 2mm},clip]{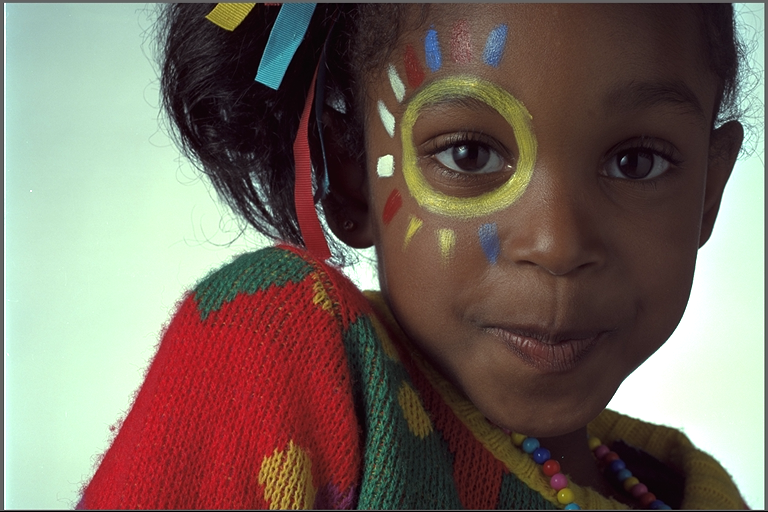}} & \\[3mm]
        \rotatebox{90}{ \hspace{7mm} \textbf{\loonie}} & \includegraphics[width=0.3\textwidth,trim={7mm 12mm 7mm 2mm},clip]{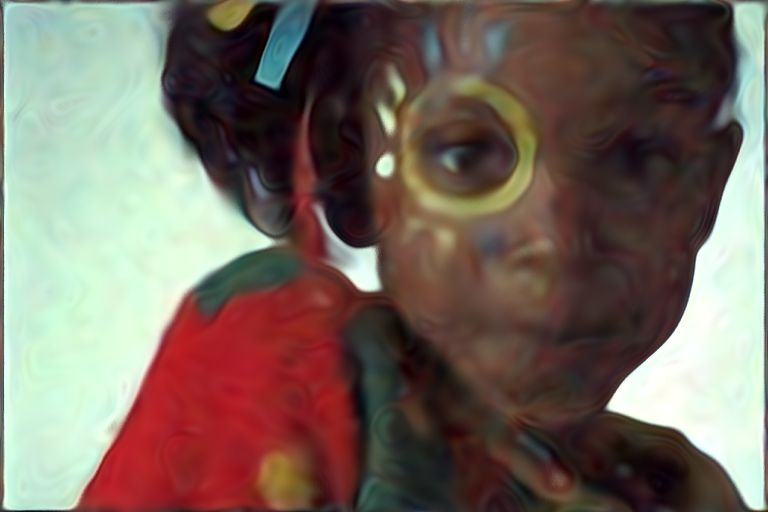} & \includegraphics[width=0.3\textwidth,trim={7mm 12mm 7mm 2mm},clip]{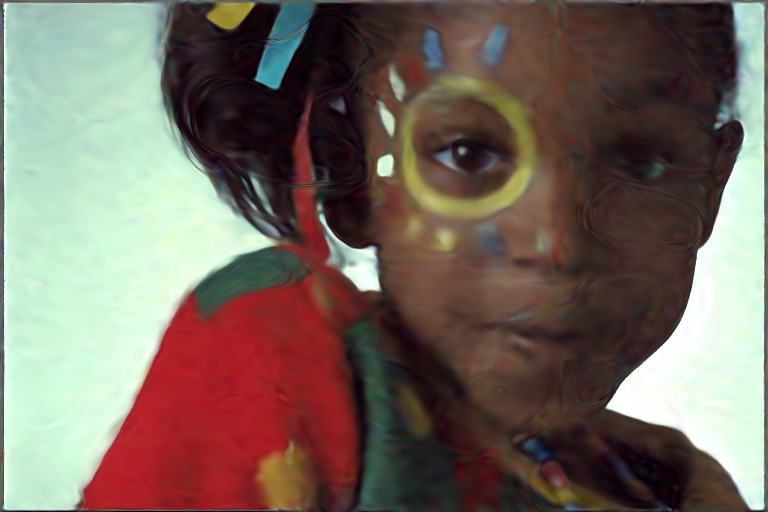} & \includegraphics[width=0.3\textwidth,trim={7mm 12mm 7mm 2mm},clip]{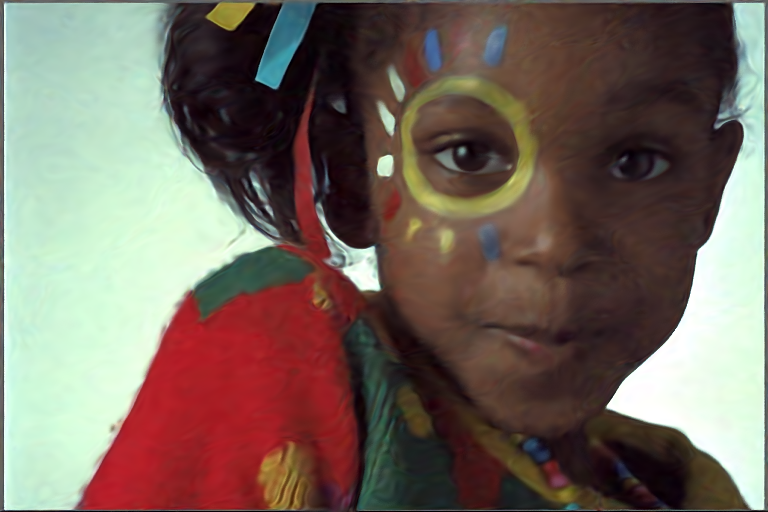} & \includegraphics[width=0.3\textwidth,trim={7mm 12mm 7mm 2mm},clip]{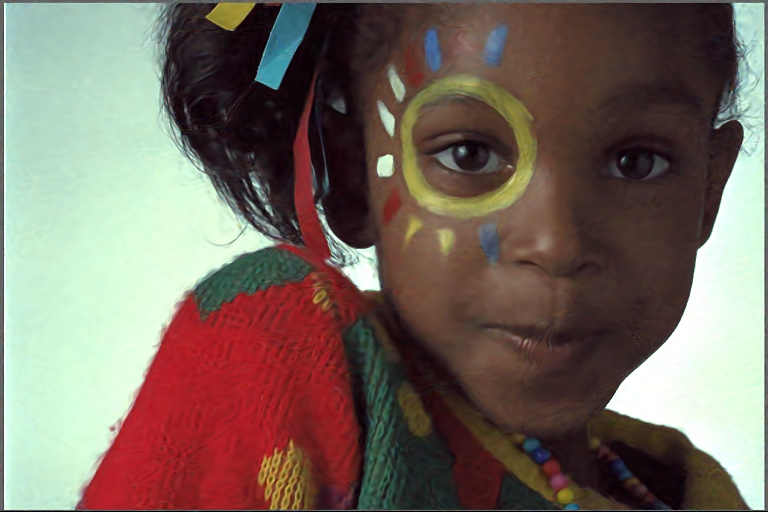} \\[3mm]
        \rotatebox{90}{ \hspace{8mm} \textbf{\coin}} & \includegraphics[width=0.3\textwidth,trim={7mm 12mm 7mm 2mm},clip]{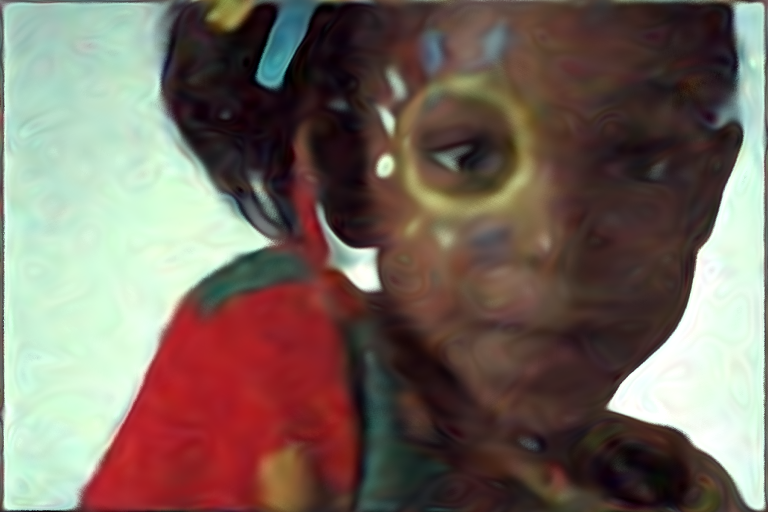} & \includegraphics[width=0.3\textwidth,trim={7mm 12mm 7mm 2mm},clip]{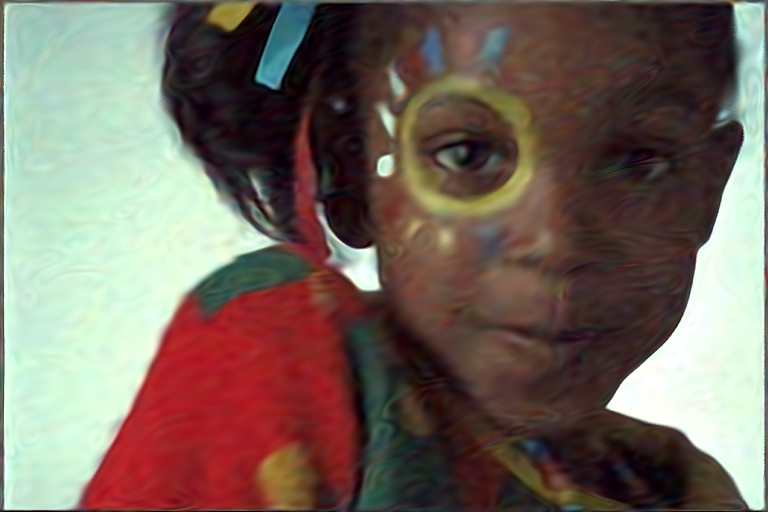} & \includegraphics[width=0.3\textwidth,trim={7mm 12mm 7mm 2mm},clip]{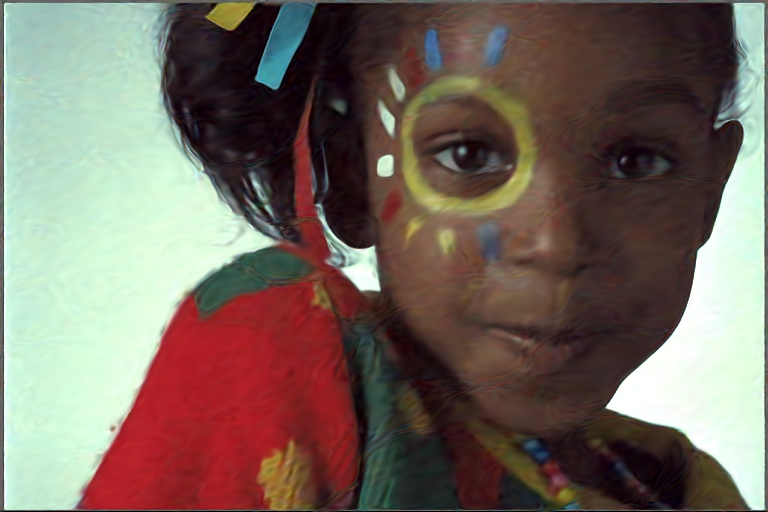} & \includegraphics[width=0.3\textwidth,trim={7mm 12mm 7mm 2mm},clip]{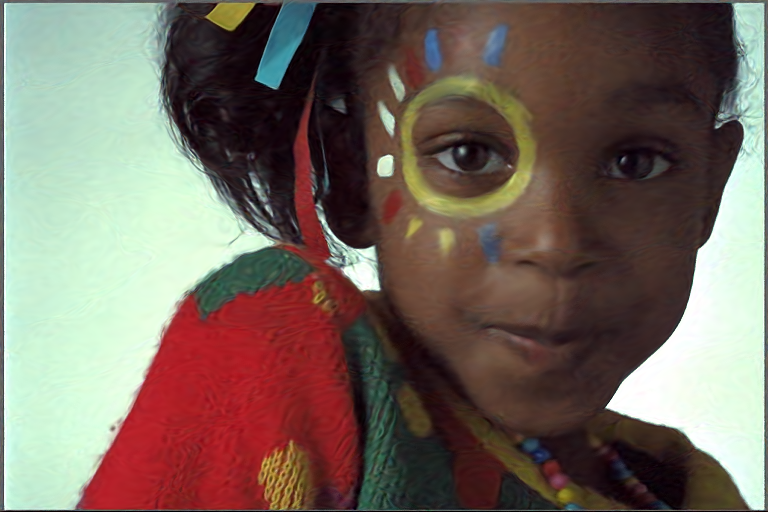} \\[3mm]
        \rotatebox{90}{ \hspace{9mm} \textbf{{\color{myred} MP}}} & \includegraphics[width=0.3\textwidth,trim={7mm 12mm 7mm 2mm},clip]{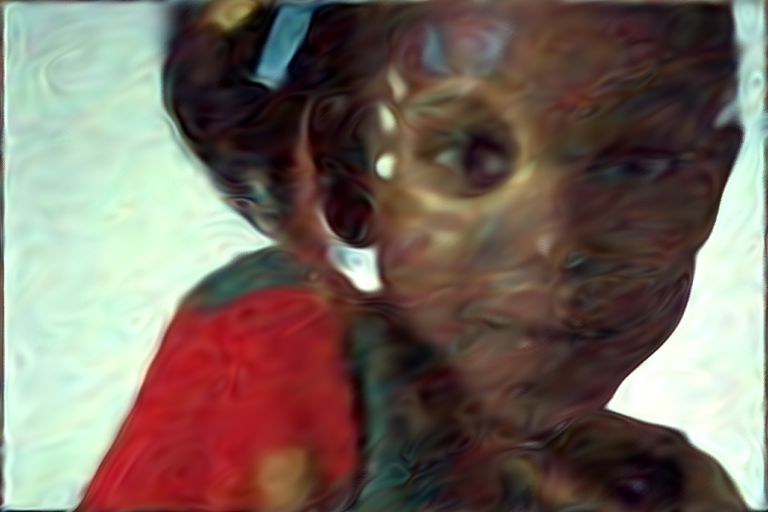} & \includegraphics[width=0.3\textwidth,trim={7mm 12mm 7mm 2mm},clip]{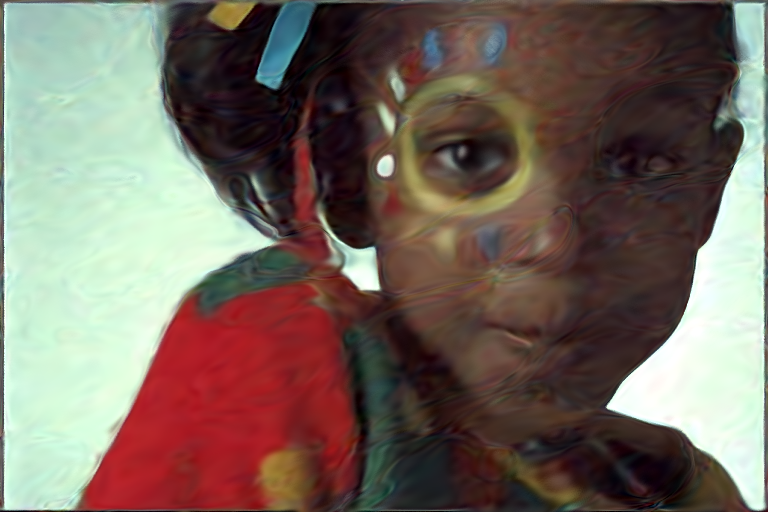} & \includegraphics[width=0.3\textwidth,trim={7mm 12mm 7mm 2mm},clip]{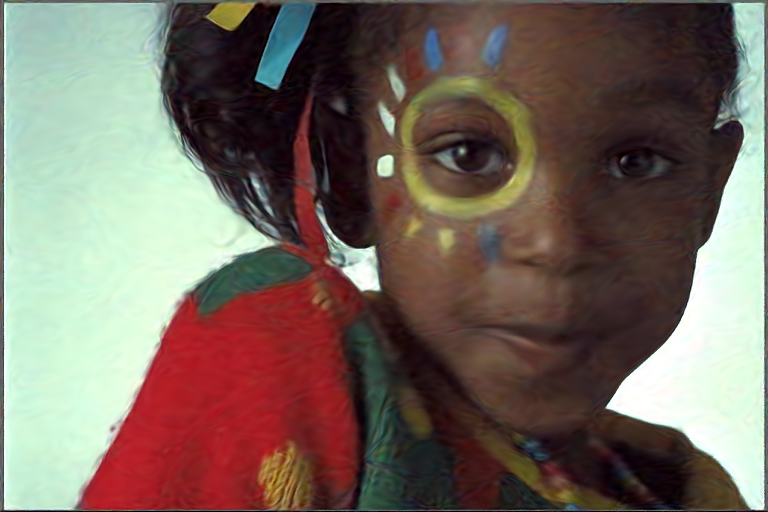} & \includegraphics[width=0.3\textwidth,trim={7mm 12mm 7mm 2mm},clip]{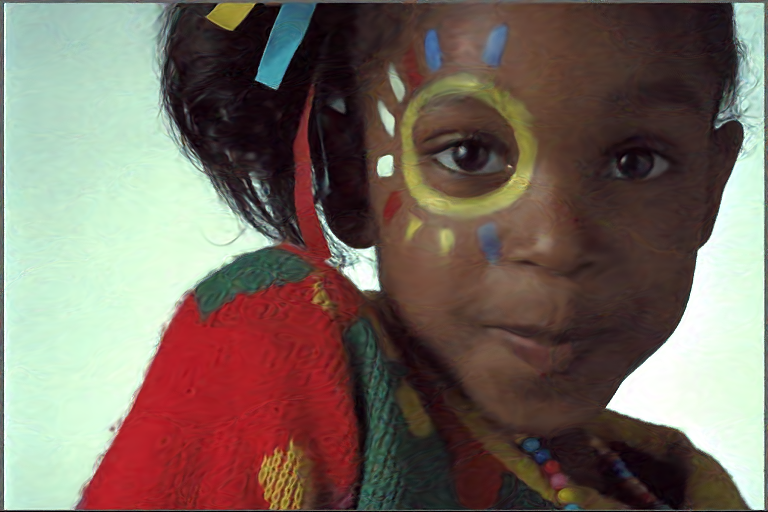} \\[3mm]
        \rotatebox{90}{ \hspace{8mm} \textbf{\jpeg}} & \includegraphics[width=0.3\textwidth,trim={7mm 12mm 7mm 2mm},clip]{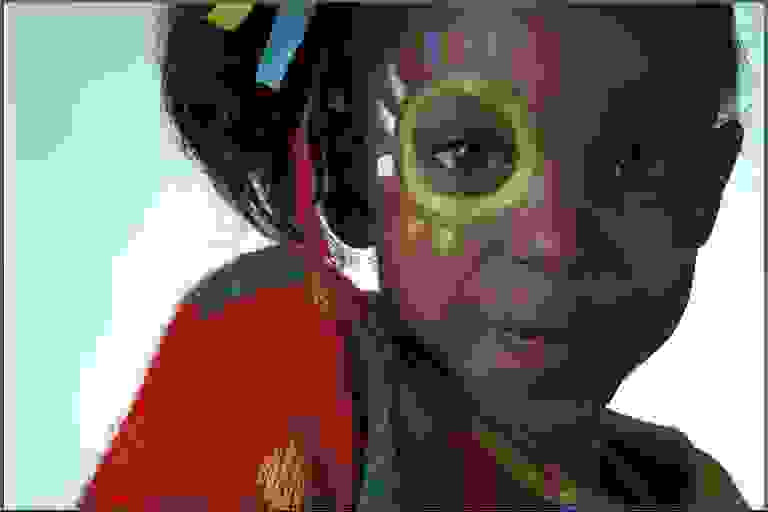} & \includegraphics[width=0.3\textwidth,trim={7mm 12mm 7mm 2mm},clip]{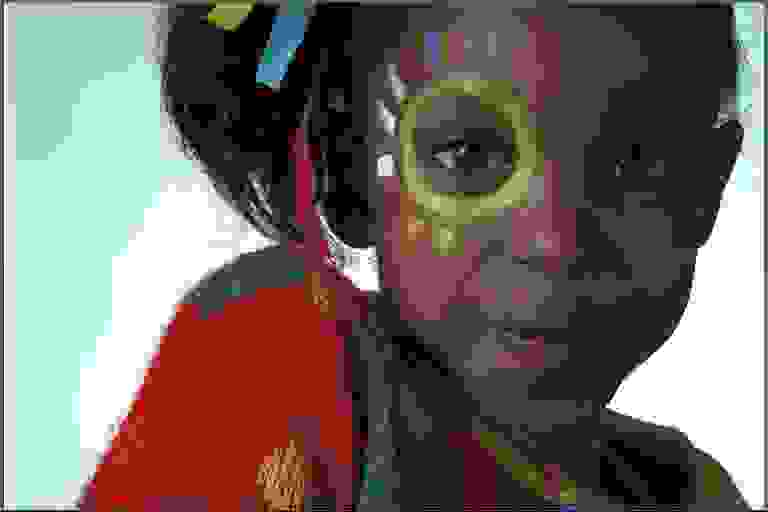} & \includegraphics[width=0.3\textwidth,trim={7mm 12mm 7mm 2mm},clip]{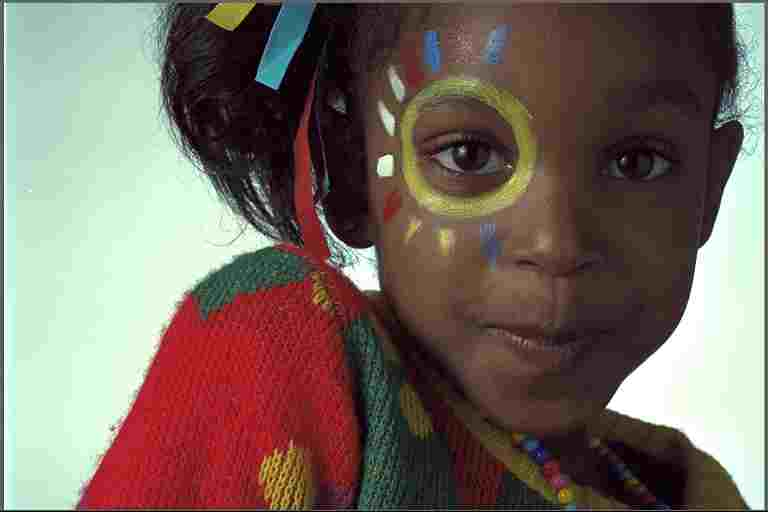} & \includegraphics[width=0.3\textwidth,trim={7mm 12mm 7mm 2mm},clip]{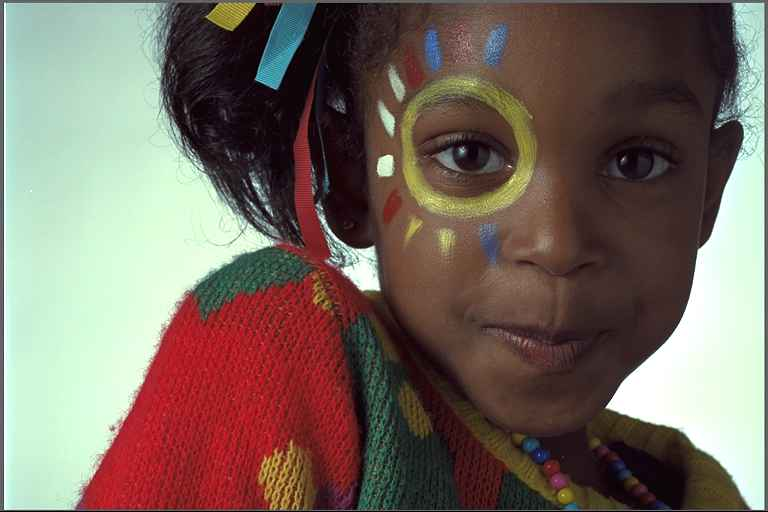} \\[3mm]
        & \textbf{0.07 BPP} & \textbf{0.15 BPP} & \textbf{0.3 BPP} & \textbf{0.6 BPP}
    \end{tabular}
    \vspace{-2mm}
    \caption{Reconstruction of Kodak image 15 for all methods at different target BPP.}
    \label{fig:kodim15}
\end{figure}

\begin{figure}[h]
    \hspace{-28mm}
    \begin{tabular}{ccccc}
        \centering
        & & \multicolumn{2}{c}{ \rotatebox{90}{\hspace{6mm} \textbf{Original}} \includegraphics[width=0.3\textwidth,trim={7mm 12mm 7mm 2mm},clip]{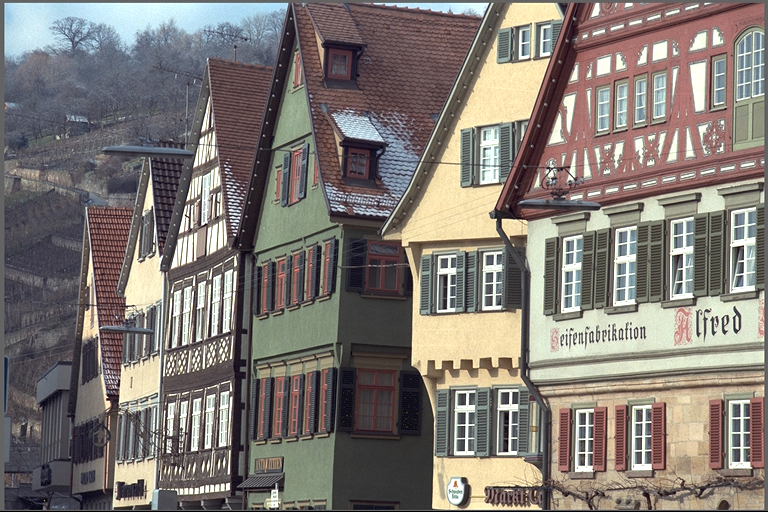}} & \\[3mm]
        \rotatebox{90}{ \hspace{7mm} \textbf{\loonie}} & \includegraphics[width=0.3\textwidth,trim={7mm 12mm 7mm 2mm},clip]{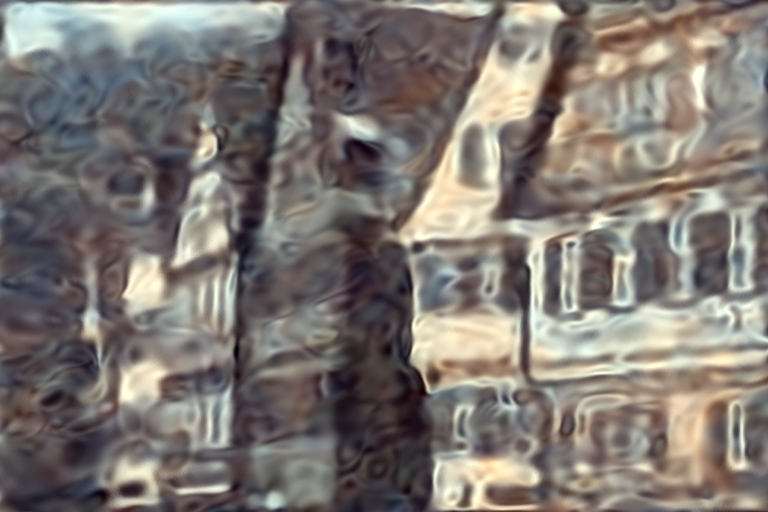} & \includegraphics[width=0.3\textwidth,trim={7mm 12mm 7mm 2mm},clip]{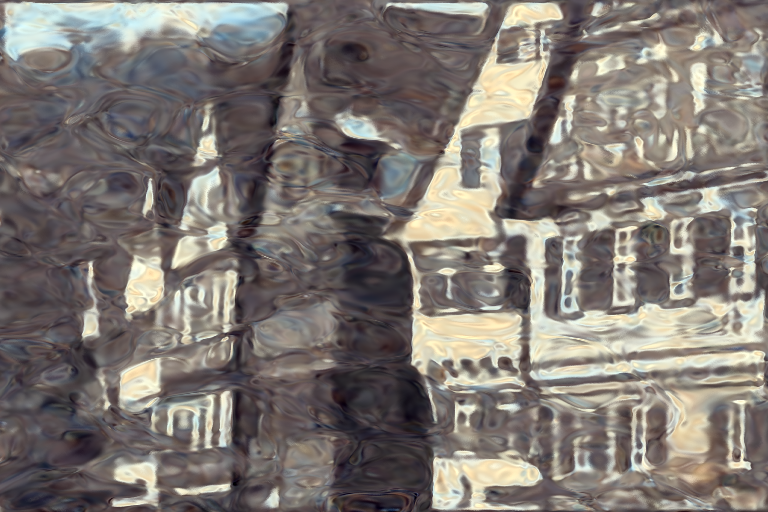} & \includegraphics[width=0.3\textwidth,trim={7mm 12mm 7mm 2mm},clip]{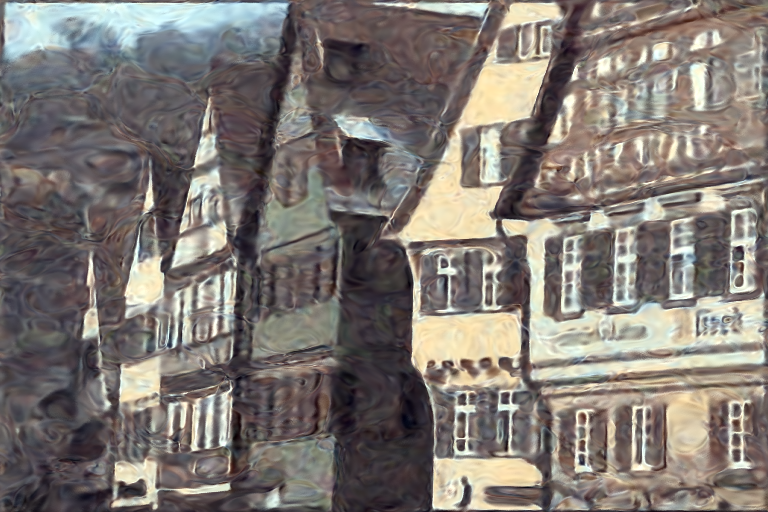} & \includegraphics[width=0.3\textwidth,trim={7mm 12mm 7mm 2mm},clip]{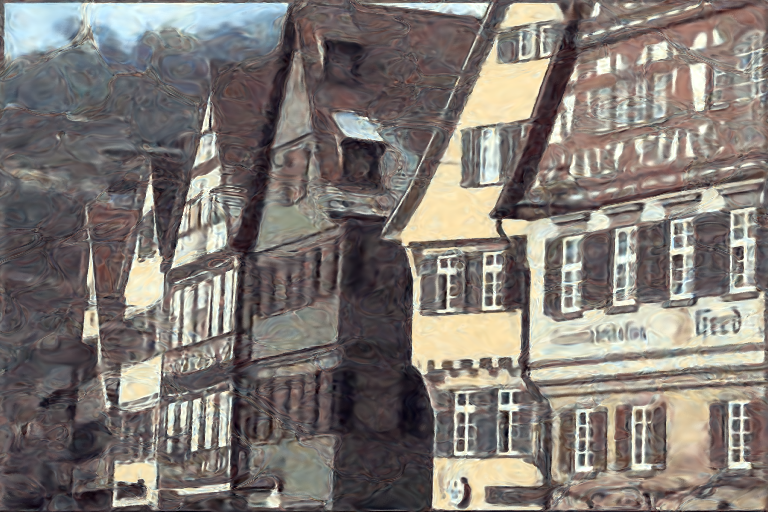} \\[3mm]
        \rotatebox{90}{ \hspace{8mm} \textbf{\coin}} & \includegraphics[width=0.3\textwidth,trim={7mm 12mm 7mm 2mm},clip]{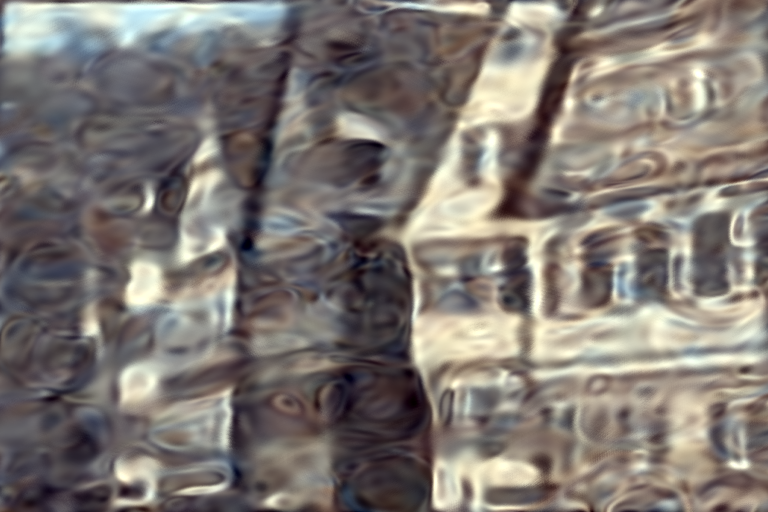} & \includegraphics[width=0.3\textwidth,trim={7mm 12mm 7mm 2mm},clip]{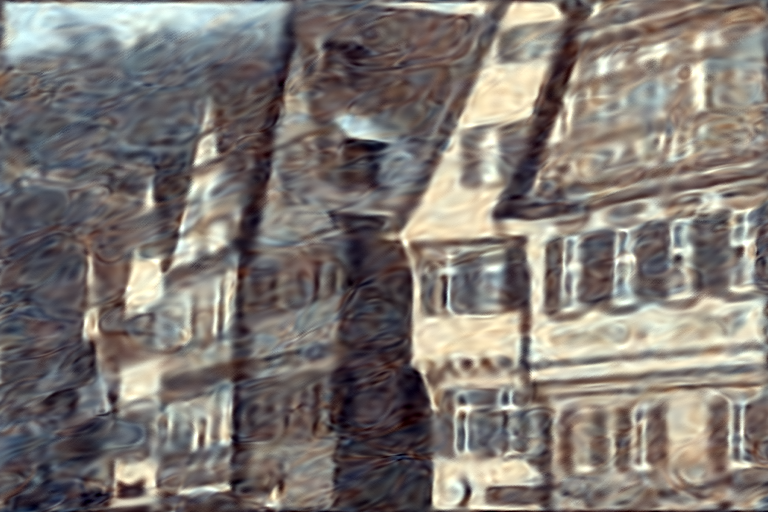} & \includegraphics[width=0.3\textwidth,trim={7mm 12mm 7mm 2mm},clip]{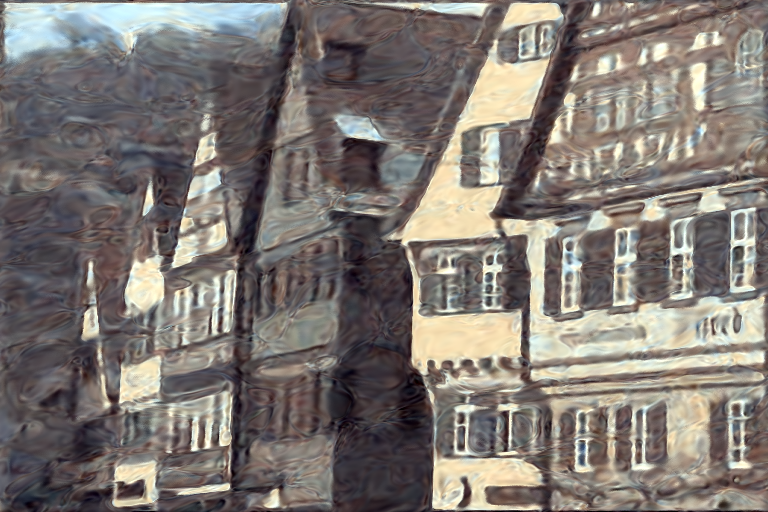} & \includegraphics[width=0.3\textwidth,trim={7mm 12mm 7mm 2mm},clip]{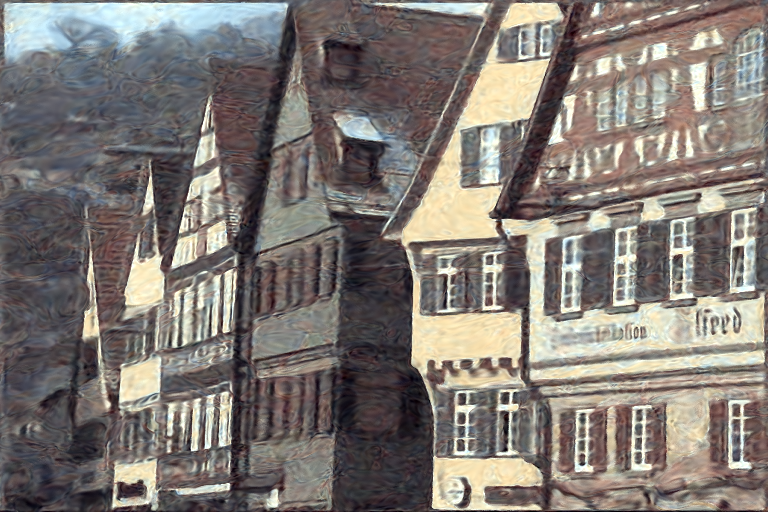} \\[3mm]
        \rotatebox{90}{ \hspace{9mm} \textbf{{\color{myred} MP}}} & \includegraphics[width=0.3\textwidth,trim={7mm 12mm 7mm 2mm},clip]{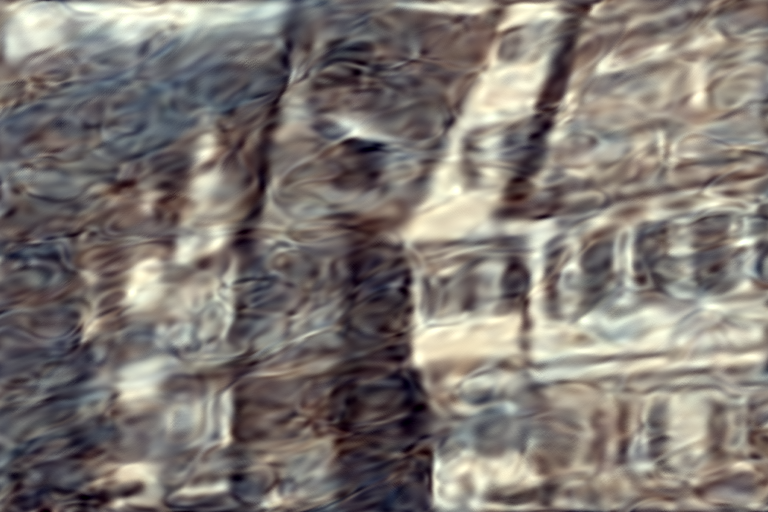} & \includegraphics[width=0.3\textwidth,trim={7mm 12mm 7mm 2mm},clip]{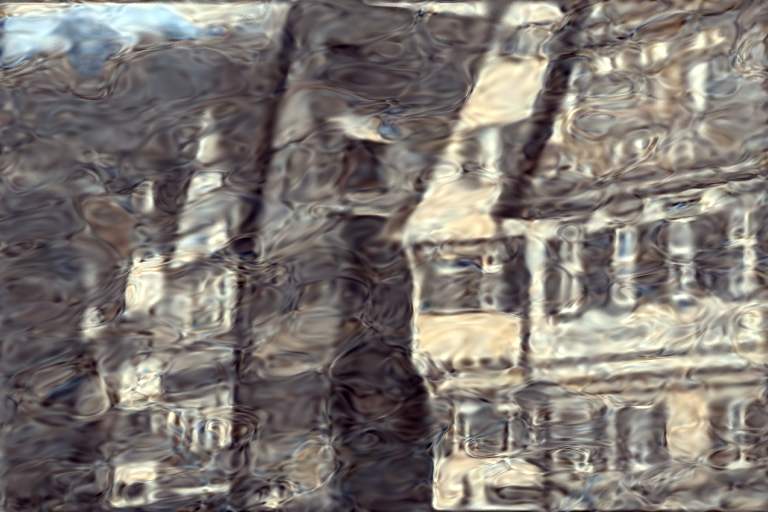} & \includegraphics[width=0.3\textwidth,trim={7mm 12mm 7mm 2mm},clip]{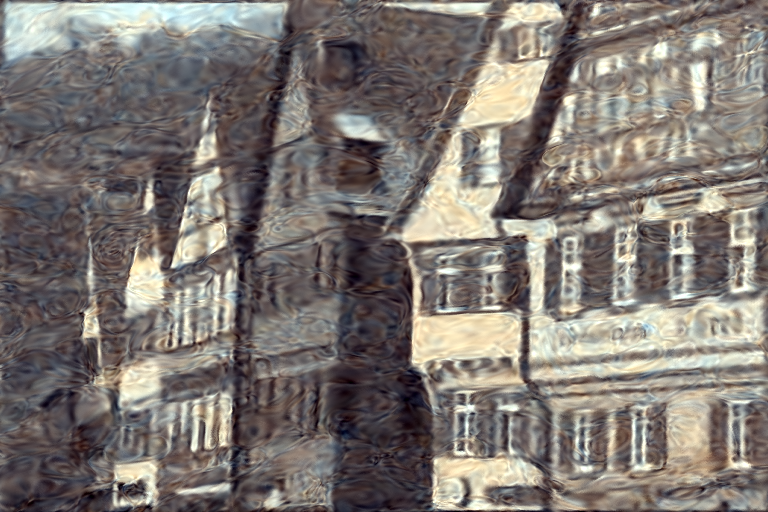} & \includegraphics[width=0.3\textwidth,trim={7mm 12mm 7mm 2mm},clip]{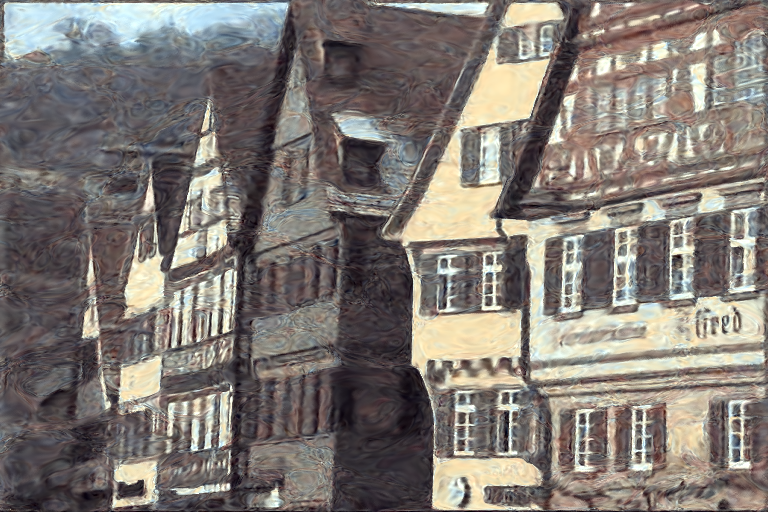} \\[3mm]
        \rotatebox{90}{ \hspace{8mm} \textbf{\jpeg}} & \includegraphics[width=0.3\textwidth,trim={7mm 12mm 7mm 2mm},clip]{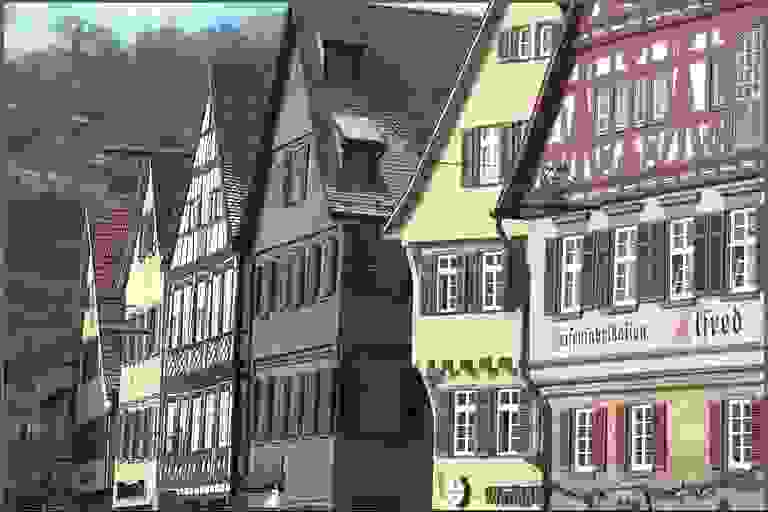} & \includegraphics[width=0.3\textwidth,trim={7mm 12mm 7mm 2mm},clip]{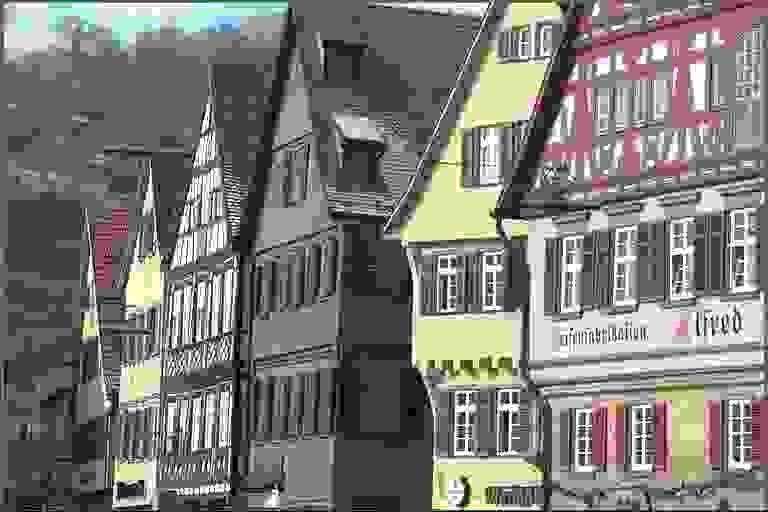} & \includegraphics[width=0.3\textwidth,trim={7mm 12mm 7mm 2mm},clip]{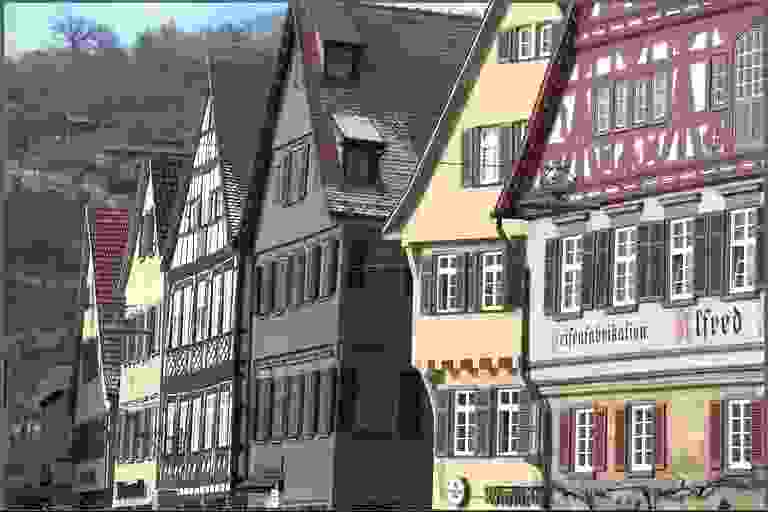} & \includegraphics[width=0.3\textwidth,trim={7mm 12mm 7mm 2mm},clip]{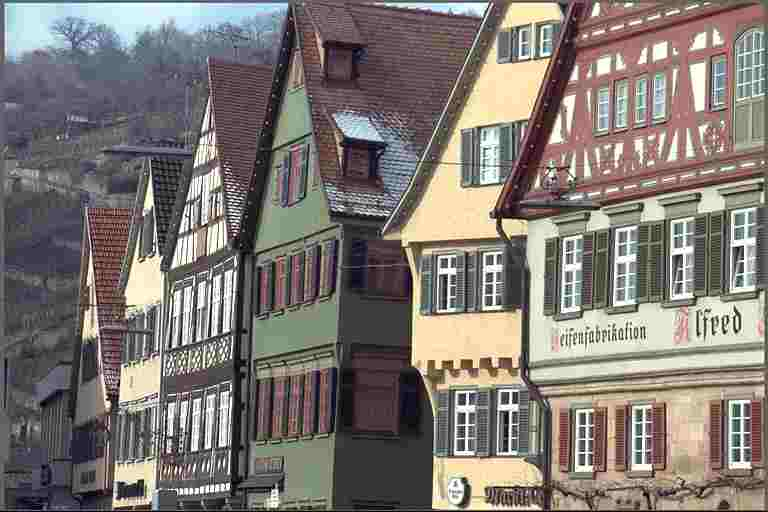} \\[3mm]
        & \textbf{0.07 BPP} & \textbf{0.15 BPP} & \textbf{0.3 BPP} & \textbf{0.6 BPP}
    \end{tabular}
    \vspace{-2mm}
    \caption{Reconstruction of Kodak image 8 for all methods at different target BPP.}
    \label{fig:kodim08}
\end{figure}

\begin{figure}[h]
    \hspace{-28mm}
    \begin{tabular}{ccccc}
        \centering
        & & \multicolumn{2}{c}{ \rotatebox{90}{\hspace{6mm} \textbf{Original}} \includegraphics[width=0.3\textwidth,trim={7mm 12mm 7mm 2mm},clip]{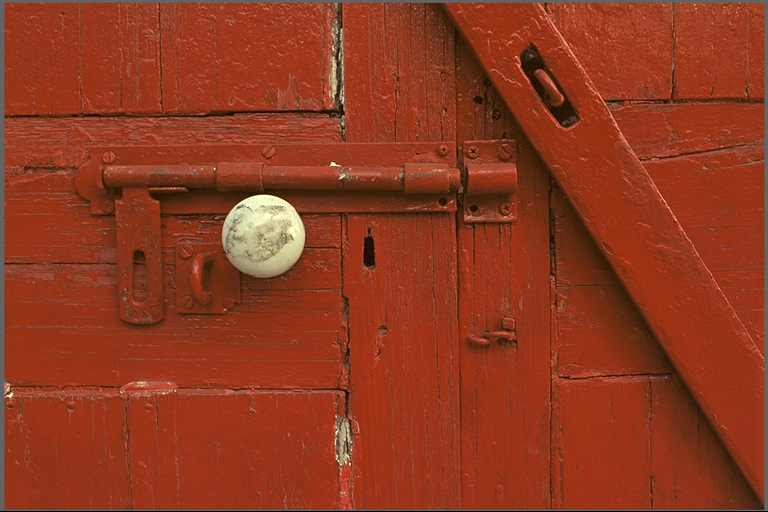}} & \\[3mm]
        \rotatebox{90}{ \hspace{7mm} \textbf{\loonie}} & \includegraphics[width=0.3\textwidth,trim={7mm 12mm 7mm 2mm},clip]{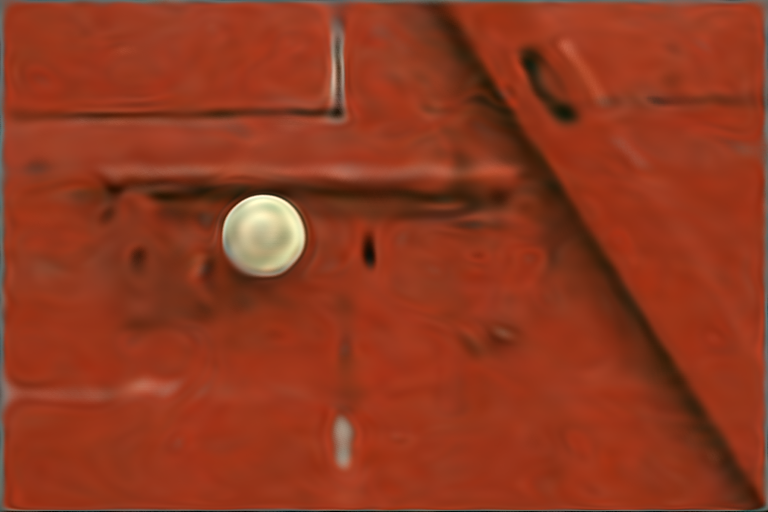} & \includegraphics[width=0.3\textwidth,trim={7mm 12mm 7mm 2mm},clip]{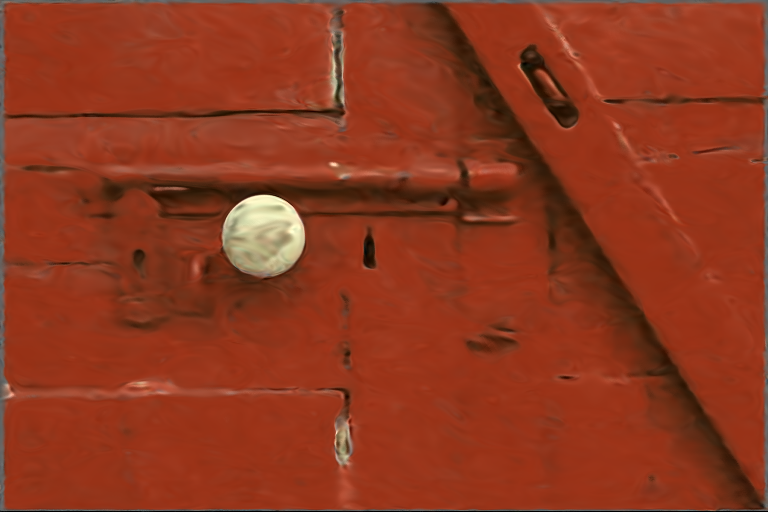} & \includegraphics[width=0.3\textwidth,trim={7mm 12mm 7mm 2mm},clip]{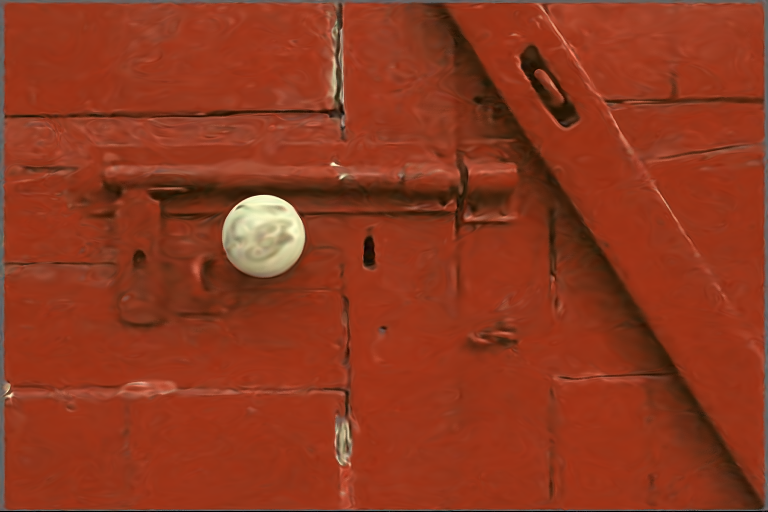} & \includegraphics[width=0.3\textwidth,trim={7mm 12mm 7mm 2mm},clip]{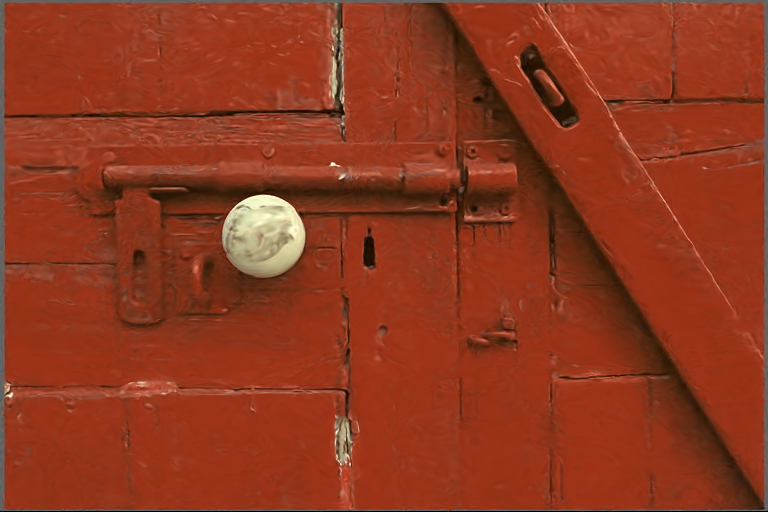} \\[3mm]
        \rotatebox{90}{ \hspace{8mm} \textbf{\coin}} & \includegraphics[width=0.3\textwidth,trim={7mm 12mm 7mm 2mm},clip]{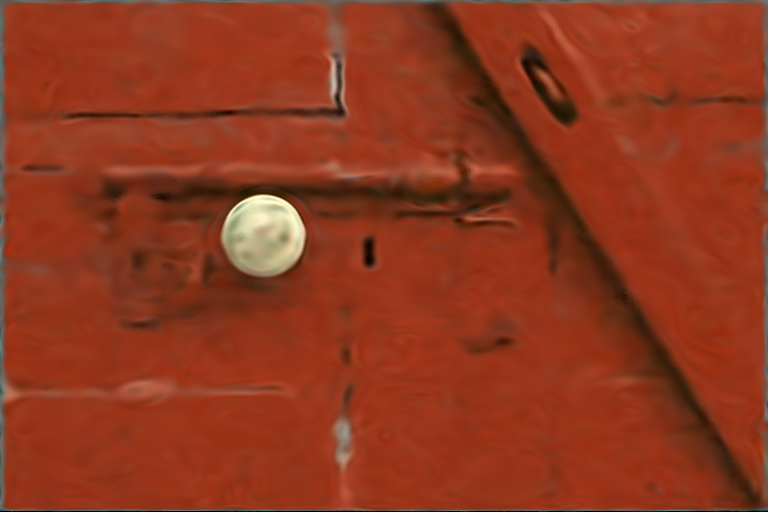} & \includegraphics[width=0.3\textwidth,trim={7mm 12mm 7mm 2mm},clip]{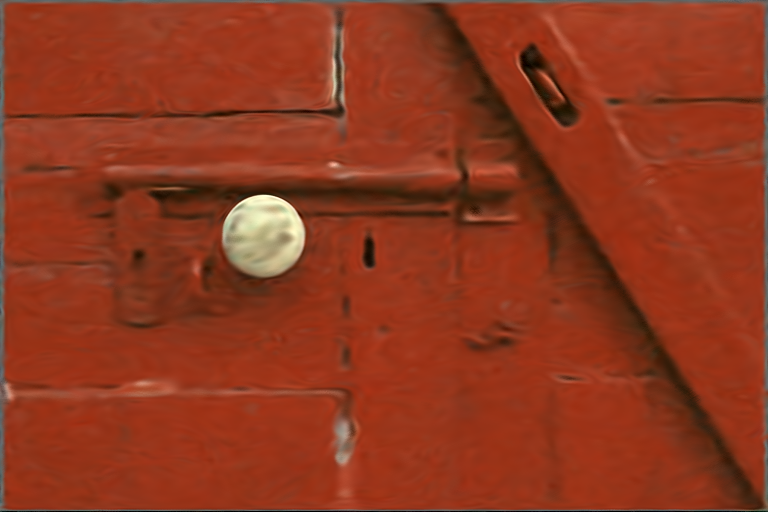} & \includegraphics[width=0.3\textwidth,trim={7mm 12mm 7mm 2mm},clip]{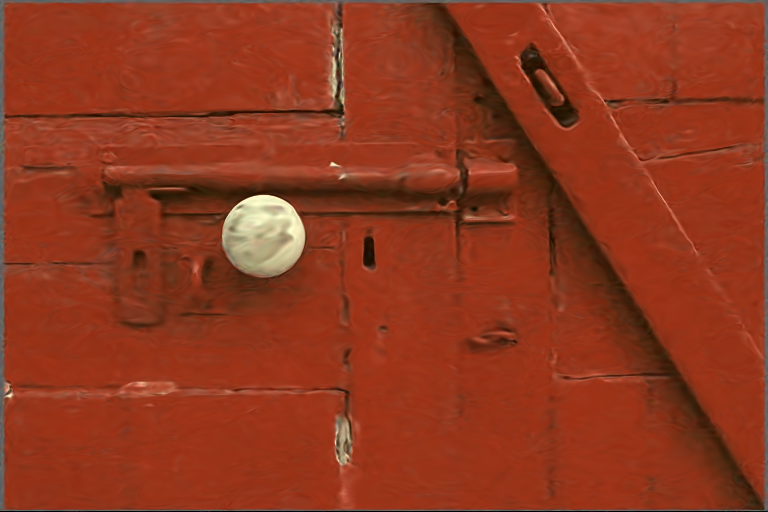} & \includegraphics[width=0.3\textwidth,trim={7mm 12mm 7mm 2mm},clip]{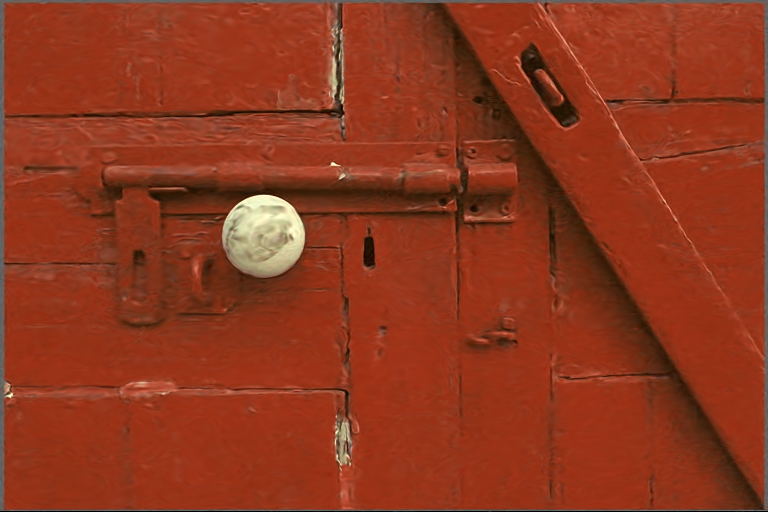} \\[3mm]
        \rotatebox{90}{ \hspace{9mm} \textbf{{\color{myred} MP}}} & \includegraphics[width=0.3\textwidth,trim={7mm 12mm 7mm 2mm},clip]{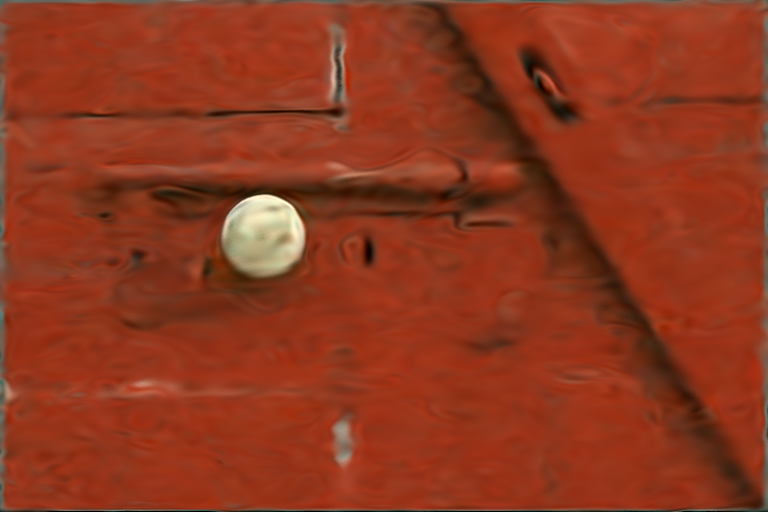} & \includegraphics[width=0.3\textwidth,trim={7mm 12mm 7mm 2mm},clip]{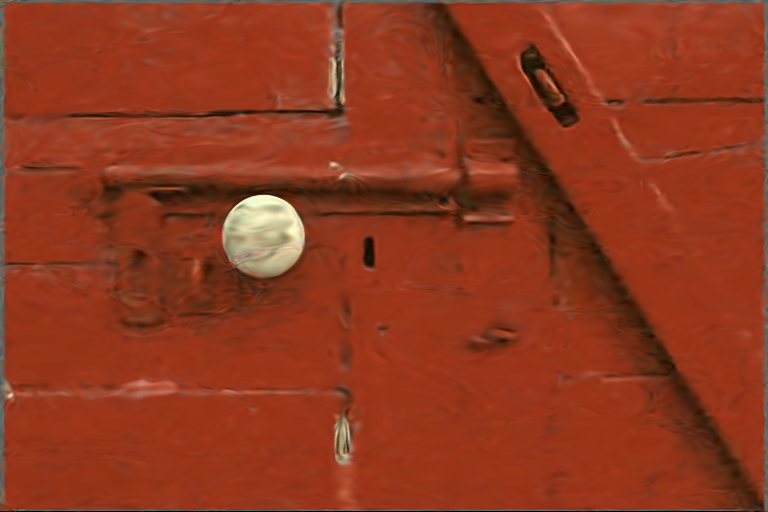} & \includegraphics[width=0.3\textwidth,trim={7mm 12mm 7mm 2mm},clip]{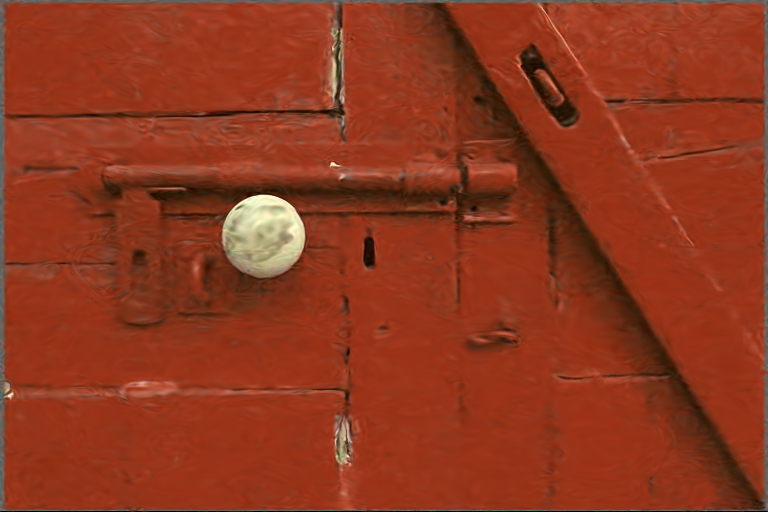} & \includegraphics[width=0.3\textwidth,trim={7mm 12mm 7mm 2mm},clip]{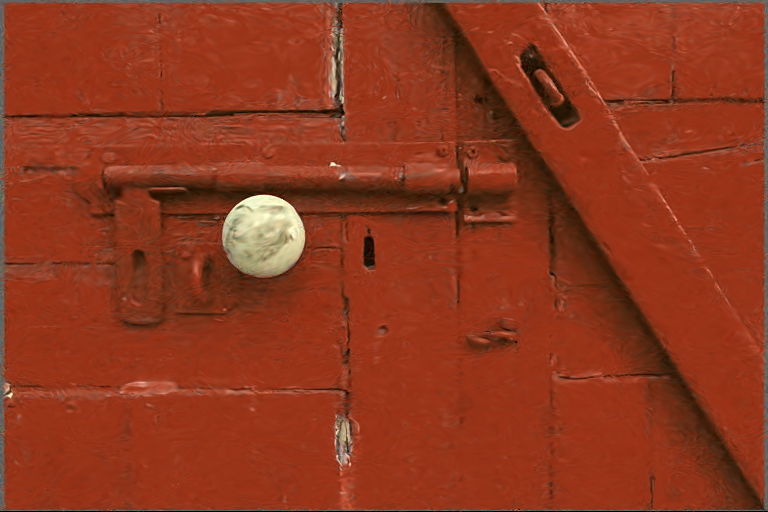} \\[3mm]
        \rotatebox{90}{ \hspace{8mm} \textbf{\jpeg}} & \includegraphics[width=0.3\textwidth,trim={7mm 12mm 7mm 2mm},clip]{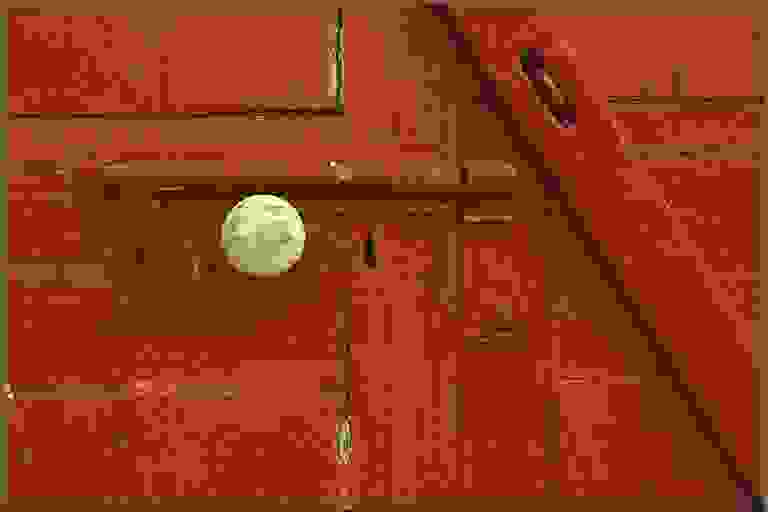} & \includegraphics[width=0.3\textwidth,trim={7mm 12mm 7mm 2mm},clip]{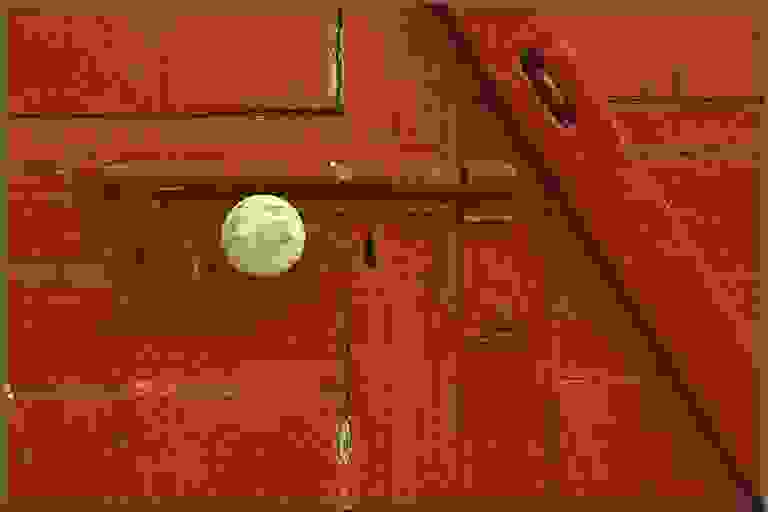} & \includegraphics[width=0.3\textwidth,trim={7mm 12mm 7mm 2mm},clip]{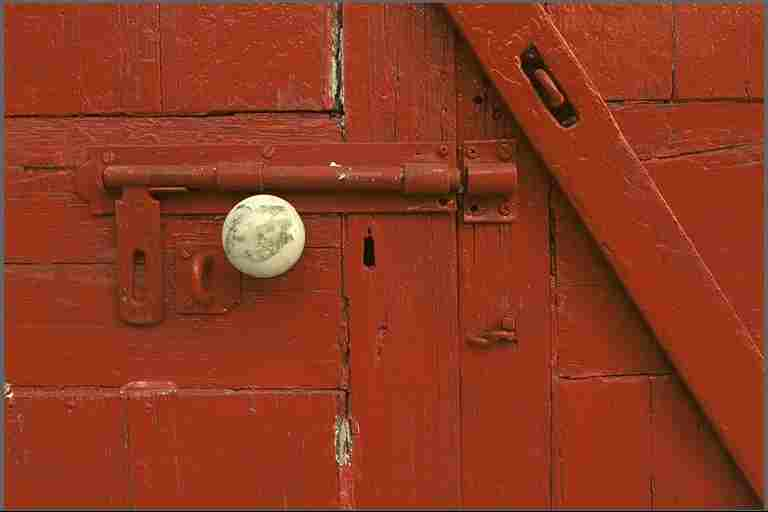} & \includegraphics[width=0.3\textwidth,trim={7mm 12mm 7mm 2mm},clip]{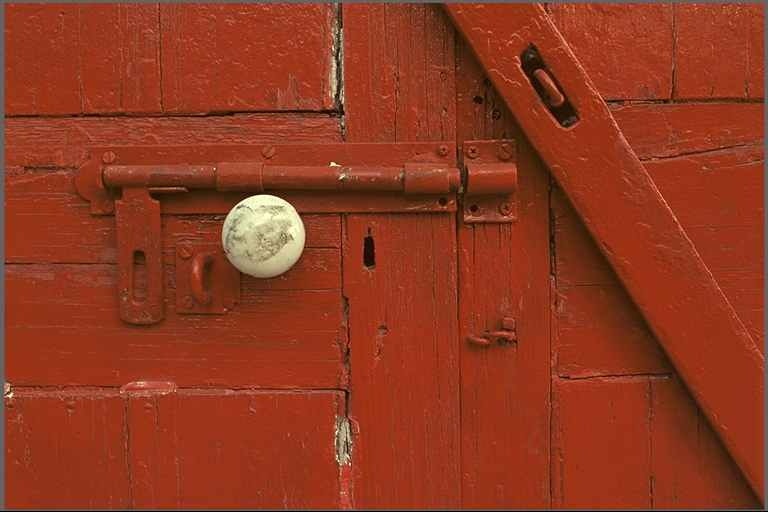} \\[3mm]
        & \textbf{0.07 BPP} & \textbf{0.15 BPP} & \textbf{0.3 BPP} & \textbf{0.6 BPP}
    \end{tabular}
    \vspace{-2mm}
    \caption{Reconstruction of Kodak image 2 for all methods at different target BPP.}
    \label{fig:kodim02}
\end{figure}

\end{document}